\documentclass[lettersize,journal]{IEEEtran}
\usepackage{amsmath,amsfonts}
\usepackage{algorithmic}
\usepackage{algorithm}
\usepackage{array}
\usepackage{textcomp}
\usepackage{stfloats}
\usepackage{url}
\usepackage{verbatim}
\usepackage{graphicx}
\usepackage{cite}
\usepackage{tabularx}
\usepackage{multirow}
\usepackage{caption}
\usepackage{subcaption}
\usepackage[dvipsnames]{xcolor}
\pdfoutput=1

\begin{document}


\title{Adaptive Segmentation-Based Initialization for Steered Mixture of Experts Image Regression}

\author{Yi-Hsin~Li, Sebastian~Knorr,~\IEEEmembership{Senior Member,~IEEE,}     Mårten~Sjöström,~\IEEEmembership{Senior Member,~IEEE,}   \\Thomas Sikora,~\IEEEmembership{Senior Member,~IEEE}\vspace{-3ex}
\thanks{Manuscript received 16 September 2024. This project has received funding from the European Union’s Horizon 2020 research and innovation program under the Marie Skłodowska-Curie grant agreement No 956770. \textit{(Corresponding authors: Thomas~Sikora.)}}
\thanks{Yi-Hsin~Li is with Department of Telecommunication Systems, Technical University of Berlin, Berlin, 10587, Germany, and also with Department of Computer and Electrical Engineering, Mid Sweden University, Sundsvall, 85170, Sweden (e-mail: yi-hsin.li@tu-berlin.de).}
\thanks{Sebastian~Knorr is with School of Computing, Communication and Business, Hochschule für Technik und Wirtschaft Berlin, Berlin, 12459, Germany (e-mail: sebastian.knorr@htw-berlin.de).}
\thanks{Mårten~Sjöström is with Department of Computer and Electrical Engineering, Mid Sweden University, Sundsvall, 85170, Sweden (e-mail: Marten.Sjostrom@miun.se).}
\thanks{Thomas~Sikora is with Department of Telecommunication Systems, Technical University of Berlin, Berlin, 10587, Germany (e-mail: thomas.sikora@tu-berlin.de).}
}



\maketitle

\begin{abstract}
Kernel image regression methods have shown to provide excellent efficiency in many image processing task, such as image and light-field compression, Gaussian Splatting, denoising and super-resolution. The estimation of parameters for these methods frequently employ gradient descent iterative optimization, which poses significant computational burden for many applications. In this paper, we introduce a novel adaptive segmentation-based initialization method targeted for optimizing Steered-Mixture-of Experts (SMoE) gating networks and Radial-Basis-Function (RBF) networks with steering kernels. The novel initialization method allocates kernels into pre-calculated image segments. The optimal number of kernels, kernel positions, and steering parameters are derived per segment in an iterative optimization and kernel sparsification procedure. The kernel information from ``local`` segments is then transferred into a ``global" initialization, ready for use in iterative optimization of SMoE, RBF, and related kernel image regression methods. Results show that drastic objective and subjective quality improvements are achievable compared to widely used regular grid initialization, ``state-of-the-art" K-Means initialization and previously introduced segmentation-based initialization methods, while also drastically improving the sparsity of the regression models. For same quality, the novel initialization results in models with around 50\% reduction of kernels. In addition, a significant reduction of convergence time is achieved, with overall run-time savings of up to 50\%. The segmentation-based initialization strategy itself admits heavy parallel computation; in theory, it may be divided into as many tasks as there are segments in the images. By accessing only four parallel GPUs, run-time savings of already 50\% for initialization are achievable.

\end{abstract}


\section{Introduction}
\IEEEPARstart{I}{mage} and video processing and computer vision have gained enormous attention in recent years, across a divers range of application domains and application specific tasks. This includes compression of digital imagery for storage and transmission, from low-rate to picture-perfect representations. Provisions for flexible resizing of images and video to provide super-resolution representations are required in many application domains to fit the imagery to the size of display or viewing conditions, e.g. for medical imaging and entertainment. Often patterns in images have to be reconstructed from noisy sensors, such as from Magnetic Resonance Spectroscopy MRT or Ultrasound images. The design of domain-specific denoising algorithms is of vital importance.

Deep Learning (DL) algorithms have made a significant impact in image processing and computer vision domains. DL strategies have shown to provide high efficiency and effectiveness in many applications, including image compression \cite{li_gan_2020,bidwe_deep_2022,yu_backdoor_2023}, object detection \cite{zhou_yolo-nl_2024,su_mod-yolo_2024}, semantic segmentation \cite{singh_deep_2024}, super-resolution \cite{lim_enhanced_2017}, inpainting \cite{yildirim_diverse_2023,zhou_propainter_2023,zhang_towards_2023}, and denoising \cite{goceri_evaluation_2023,yuan_hcformer_2023}. Despite the remarkable achievements of DL across various applications, certain drawbacks persist. Deep Neural Network models are often characterized by their complexity and depth of architectural design, hindering their scalability. For instance, in super-resolution, sophisticated DL models \cite{lim_enhanced_2017} may struggle with scalability due to their intricate architectures. Often optimized to provide an excellent representation of images at a certain level of magnification, they cannot work on other levels or they significantly underperform. Furthermore, lack of interpretability can limit their application in critical domains where model transparency is essential, e.g. in medical imaging. 

In contrast to DL, shallow networks, such as Radial Basis Function networks (RBF) \cite{ghosh_overview_2001}, Gaussian splatting \cite{kerbl_3d_2023}, Mixture-of-Experts networks or Steered-Mixture-of-Experts Gating networks \cite{verhack_steered_2020,verhack_universal_2016,jongebloed_sparse_2022,jongebloed_quantized_2019,bochinski_regularized_2018,fleig_edge-aware_2022,fleig_steered_2023}, often offer advantages such as high interpretability of results, adaptability and a unified framework across several processing tasks. Steered Mixture-of-Experts (SMoE) models are sparse gating networks that have emerged in recent years in applications such as image, video and light-field representation and compression \cite{jongebloed_sparse_2022,fleig_edge-aware_2022}, super-resolution \cite{Ozkan_Edge_2024}, and denoising \cite{ozkan_steered_2023,Ozkan_Edge_2024, fleig_steered_2023}. SMoE networks operate as a parametric regression method, leveraging gating functions with multi-dimensional kernels, i.e. of the exponential family, including Gaussians. The gating functionality is designed to provide an edge-aware representation with an explicit and flexible edge model, capable of modeling both abrupt and smooth transitions in image signals. Unlike DL models, SMoE networks rely on a mixture of specialized experts that contribute to the representation.  The transparent impact of the individual SMoE gates on the final result offers interpretability, allowing a clear understanding of the regression made by individual experts. 

The prime purpose of this paper is to address the high computational burden frequently associated with the optimization of SMoE network parameters. When iterative strategies, such as Expectation-Maximization (EM) or Gradient Descent (GD), are employed for parameter estimation, the initialization of the parameters \cite{bochinski_regularized_2018,verhack_steered_2020} prior to optimization plays a vital role for speed of convergence, quality of reconstruction and run-time of the algorithms. However, initialization remains an under-explored facet in current research, especially for SMoE gating networks.  In this paper, we investigate a novel adaptive segmentation-based (AS) initialization strategy to boost the performance of SMoE models and related kernel image regression methods.  
The contributions of the proposed initialization are three-fold:
\begin{itemize}
    \item[-]
    Adaptive Initialization for High-Frequency Detail Capture: Our proposed novel use of image segmentation locally distributes adaptive numbers of kernels into the images for initialization. This innovation enables our method to achieve superior performance in capturing high-frequency details by tailoring initialization parameters locally.
    \item[-]
    Reduced Optimization Time: Significantly shortens optimization time due to the use of well-initialized kernels, including both their number and parameters. Ensures that the initial kernels are closer to the final optimized results, reducing the time needed to achieve the final optimization.
    \item[-]
    Broad Applicability: While designed for SMoE gating model initialization, the method can also be applied to efficiently initialize other kernel regression models, such as Radial Basis Function (RBF) and ``normalized" RBF networks.
\end{itemize}

The paper is organized as follows: In Section II we provide an overview of the SMoE gating network approach and develop an understanding of the soft-gating nature of the algorithm. Particular emphasis is placed on the relevance of number and placement of the underlying Steering Kernels to achieve an edge-aware representation, tailored to the particularities of varying image content. Section III provides a review of existing initialization methods frequently employed for SMoE or related algorithms and discusses prior attempts to employ segmentation-based initialization. Section IV motivates the novel region-adaptive, segmentation-based initialization approach. The underlying two-stage process with ``Segmentation Reconstruction" and ``Optimization from Partial Parameters" is introduced. Section V and VI provide experimental setting and validation, respectively, and Section VII finally summarizes and concludes the paper.


\begin{figure}
\captionsetup{font=small}
    \centering
    \begin{subfigure}[t]{0.25\linewidth}
        \includegraphics[width=\linewidth]{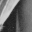}
        \caption{Original}
    \end{subfigure}
    \hfill
    \begin{subfigure}[t]{0.25\linewidth}
        \includegraphics[width=\linewidth]{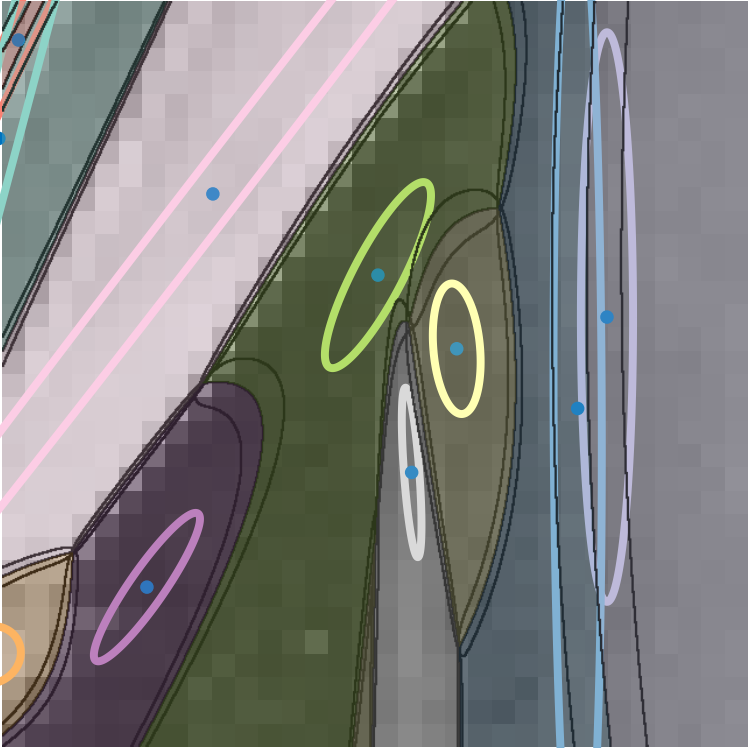}
        \caption{SMoE Kernels}
        \label{fig:1_kernels}
    \end{subfigure}%
    \hfill
    \begin{subfigure}[t]{0.25\linewidth}
        \includegraphics[width=\linewidth]{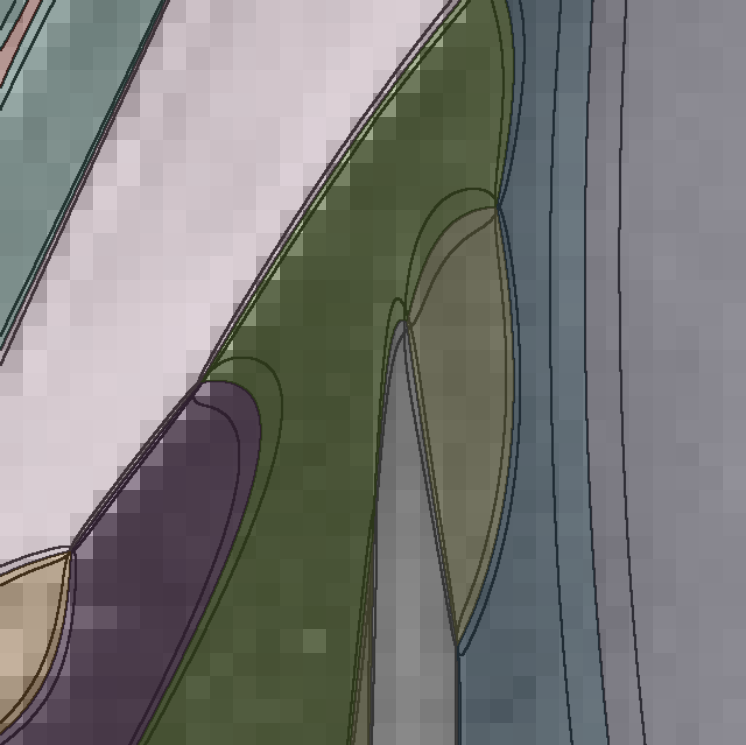}
        \caption{SMoE Gates}
        \label{fig:1_Gates}
    \end{subfigure}
    \hfill
    \begin{subfigure}[t]{0.24\linewidth}
        \includegraphics[width=\linewidth]{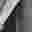}
        \caption{\textbf{JPEG}\\ PSNR:$26.33 dB$\\ SSIM: $0.82$}
        \label{fig:1_JPEG}
    \end{subfigure}%
    \hfill
    \begin{subfigure}[t]{0.24\linewidth}
        \includegraphics[width=\linewidth]{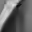}
        \caption{\textbf{HEVC}\\PSNR:$26.05 dB$ \\SSIM: $0.77$}
        \label{fig:1_HEVC}
    \end{subfigure}%
    \hfill
    \begin{subfigure}[t]{0.24\linewidth}
        \includegraphics[width=\linewidth]{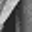}
        \caption{\textbf{JPEG2000}\\ PSNR:$29.43 dB$\\ SSIM: $0.87$}
        \label{fig:1_JPEG200}
    \end{subfigure}%
    \hfill
    \begin{subfigure}[t]{0.24\linewidth}
        \includegraphics[width=\linewidth]{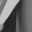}
        \caption{\textbf{SMoE}\\ PSNR:$31.66 dB$ \\SSIM: $0.9$}
        \label{fig:1_smoe}
    \end{subfigure}%
\vspace{-0.15cm}
\caption{Illustration of edge-aware Steered Mixture of Experts (SMoE) modelling for compression and denoising at 0.43 bpp (bits per pixel).}
\label{fig1}
\vspace{-0.4cm}
\end{figure}

\section{Steered Mixture of Experts (SMoE)}

While this paper focuses on initialization of GD optimization of Steered Mixture-of-Experts (SMoE) models, it is beneficial to develop a basic understanding of the definition, functioning and capabilities of the SMoE gating networks. The SMoE image model describes an edge-aware, parametric, continuous non-linear regression function. We use Fig.  \ref{fig1} (from \cite{fleig_edge-aware_2022}) to illustrate the kernel model concept for typical image compression and denoising tasks. SMoE gating networks are able to explicitly model and reconstruct sharp and smooth transitions in images without straddling edges. The sparse, edge-aware SMoE model is able to reconstruct the original image (Fig.  \ref{fig1}a) with excellent edge quality as illustrated in Fig.  \ref{fig1}g. In contrast to JPEG, JPEG2000 and HEVC-Intra at same bit rate, typical blocking and ringing artifacts are avoided completely. JPEG-like compression schemes are ``frequency"-domain techniques which quantise and code DCT- or wavelet-coefficients. This results in so-called ``mosquito" noise coding artifacts clearly seen in the reconstructed images at low rates (Fig.  \ref{fig1}d-f). Compression with SMoE models is done in the ``pixel"-domain by quantizing and coding kernel parameters. At low rates this results in geometric distortions of edges and lines, which is not directly visible in the reconstructed SMoE image. Both objective quality measures as well as subjective quality are greatly enhanced. We employ Gaussian steering kernels which are distributed into the pixel raster (Fig.  \ref{fig1}b). The position and steering parameters of the kernels are optimized using GD optimization prior to coding. Several ``experts" collaborate to explain the data in the particular 2D image region, while the associated 2D soft-gating functions (Fig.  \ref{fig1}c) define the actual influence of each expert for each pixel. In our work, the ``gates", represented by 2D softmax functions, define boundary transitions in images, such as sharp edges and smooth transitions - providing for the edge-awareness of the model. Sharp edges are modelled with sharp gating functions while smooth transitions are modelled using overlapping gates. In Fig. \ref{fig1} the sparse model with only 10 kernels is sufficient to explain the important pixel variations in the image, while ``denoising" the pixels simultaneously. 

The sparse, edge-aware representation of SMoE kernels appears to be attractive for a number of reasons for compression, denoising and beyond:
\begin{itemize}
\item[-] 
SMoEs easily extend to N-dimensional signals, including video, light-fields and light-field videos.
\item[-] 
Modeling is not restricted to regular N-dimensional pixel grids, but can be applied to irregularly sampled imagery including arbitrarily shaped segments and point clouds of any dimension.
\item[-]  The sparse (coded) representation allows to easily extract N-dimensional image features from the (decoded) model, such as edges, intensity flow, etc.
\item[-]  Once built, the continuous SMoE model allows to resample to any resolution in time and space, which includes edge-aware super-resolution and motion-interpolation.
\end{itemize}

The functioning of SMoE gating networks is significantly different compared to conventional ``kernel regression" methods such as RBF networks \cite{ghosh_overview_2001}. RBFs define regression functions $y_p(x)$ based on a ``weighted sum of $L$ kernels": 

\begin{equation}
y_p(x)={\textstyle\sum}_{j=1}^{L} m_j \cdot K_j(x).
\label{eq:1}
\end{equation}
Here $x$ is the 2-dimensional continuous signal space over which the pixels are defined and $m_j$ are the weights of the kernels $K_j(x)$. 
In stark contrast, a SMoE gating network approach operates on the ``weighted sum of $L$ soft-gates" $w_j(x)$. The regression function for SMoE gating networks is  defined as follows: 
\begin{equation}
y_p(x)={\textstyle\sum}_{j=1}^{L} m_j(x) \cdot w_j(x)
\label{eq:2}
\end{equation}
with steering Gaussian kernels
\begin{equation}
K_j(x)=\exp(-\frac{1}{2}(x-\mu_j)^T\Sigma_j^{-1}(x-\mu_j))
\label{eq:3}
\end{equation}
and associated gating functions $w_j(x)$, defined by softmax functions of the Gaussian kernels
\begin{equation}
w_j (x) = \frac{\pi_j \cdot K_j(x)}{{\textstyle\sum}_{i=1}^{L}\pi_i \cdot K_i(x)} .
\label{eq:4}
\end{equation}
$\mu_j$ and $\Sigma_j$ are the center vectors and the covariance matrices of the steered Gaussian kernels, respectively. 

$m_j(x)$ are the so-called expert functions which can be of any functional form, including constant, linear, quadratic as well as DCT and wavelet bases. In our work, we employ simple constant experts $m_j(x)=m_j$. Note that for grayscale images, a constant value suffices, while for RGB images, this constant value can be easily extended to a 3-dimensional vector, representing the corresponding RGB color values. This representation has shown excellent results in previous work \cite{tok_mse_2018}. Notice, that the SMoE regression model can be seen as an extension of the ``normalized" radial basis function network \cite{heimes_normalized_1998}, yet with steered kernels. According to Eq. (\ref{eq:4}), irregularly shaped 2D softmax gating functions $w_j (x)$  in Fig. \ref{fig1}c are derived based on the position and parameters of the $L$ kernels in Fig. \ref{fig1}b. 


Both RBF as well as SMoE model parameters are estimated in an optimization process. From Fig. \ref{fig1} and the regression function above, it is apparent that the accuracy of the estimated parameters of the Gaussian SMoE kernels is of vital importance for deriving accurate gating functions -- that allow for precise reconstruction of edges, lines, and smooth transitions -- in short, for reconstructing the infinite variety of textures in images with sufficient quality. Obviously, also the number of kernels $L$ is of major impact. The prevalent training approach for Gaussian Mixture Models (GMMs) and Mixture of Experts (MoEs) is the EM algorithm \cite{jank_em_2006}. However, EM optimizes the likelihood of a mixture model with respect to a given pixel location $x$. EM optimization was used in earlier works on SMoE regression \cite{verhack_steered_2020}. In regression tasks, optimizing an underlying model directly by Gradient Descent (GD) \cite{ruder_overview_2017} with a Mean Squared Error (MSE) objective function results in significantly improved models. Rather than using the EM likelihood as the quality metric, GD optimization based on the MSE between image data $y(x)$ and parametric regression $y_p(x)$ provides a more sensible optimization criterion. In addition, a GD approach allows the use of more sophisticated loss functions such as Multi-Scale Structural Index Measure (MS-SSIM) \cite{wang_multiscale_2003} and PSNR \cite{hore_image_2010}, as well as regularization and kernel sparsification strategies \cite{bochinski_regularized_2018} to improve the results. GD is highly effective in achieving high-quality results through iterative parameter adjustments. Despite its effectiveness, the iterative nature of GD, where the algorithm incrementally adjusts parameters in the direction of the steepest descent, demands extensive computational resources. This high computational demand can be a bottleneck, particularly when dealing with large datasets or complex models. As a result, the efficiency of GD becomes a critical consideration. In scenarios where computational resources are constrained, its computational expense can hinder the practicality of its application. 

Accelerating GD optimization in the global SMoE paradigm with its significantly improved
results remains a crucial challenge, and research efforts are needed to devise strategies that efficiently exploit the inherent parallelism within global optimization without sacrificing nuanced representation.

\section{Initialization of Parameters}

Initialization plays a pivotal role in accelerating the optimization process for global SMoE or RBF models \cite{verhack_steered_2020,bochinski_regularized_2018}. Good initialization of kernel parameters for GD optimization results in faster convergence and improved loss. This is especially the case for the highly non-linear SMoE regression problems, where the gating networks are trained using optimization techniques on non-convex domains. 

Well-known kernel methods such as RBF Networks \cite{ghosh_overview_2001} and Kernel Regression \cite{takeda_kernel_2007} or even Support Vector Regression \cite{drucker_support_1996} work on the ``weighted sum of kernels". These methods require many kernels to support complex patterns in images, in particular sharp edges. The sparsity of these methods is limited. On the other hand, it is often relatively straight-forward to understand the impact of each kernel on the reconstruction of a particular pixel. In a drastic departure, SMoE models require to compute the ``weighted sum of soft-gates", whereby each kernel contributes to the definition of each gate. The influence of each kernel and gate can be potentially of ``global" nature. Kernels may have an impact over the entire image space - and may support the reconstruction of extremely remote pixels. This ``global" functionality is in fact the desired attribute of SMoEs for many applications, because it allows to derive extremely sparse SMoE models. Single kernels may harvest the correlation over hundreds of pixels which is an essential requirement for applications such as compression and denoising \cite{jongebloed_quantized_2019},\cite{jongebloed_sparse_2022}. On the other hand, it is extremely challenging to find a good SMoE initialization strategy that helps to avoid poor local minima and to accelerate convergence. To decide on the number of kernels and especially, where to place the kernels in an initialization stage is by far less straight-forward compared to traditional ``sum of kernels" methods.

Global-based SMoE gating networks typically involve many kernels and corresponding experts, each with its own set of parameters. Ensuring diversity among these components during initialization is crucial to prevent them from collapsing into similar solutions and to encourage exploration of the input space. A good number of studies have introduced diverse SMoE initialization methodologies, each offering unique insights and advantages.

\subsection{Hierarchical-based Initialization}
Hierarchical-based initialization provides a systematic way to model complex distributions. Jongebloed et al. \cite{jongebloed_hierarchical_2018} focused on a hierarchical method that employs a tree-structured splitting strategy to determine the number of kernels. This involves analyzing the density distribution of the data and adapting the number of kernels accordingly. A hierarchical clustering structure is adopted to guide the construction of the GMM components, involving splitting kernels based on training loss. Kerbl et al. \cite{kerbl_3d_2023} introduced a merging and splitting densification scheme to control the total number of kernels. This method provides a data-driven, adaptive structure, enhancing the model's capacity to capture diverse patterns in the data. A draw-back, depending on the depth of the hierarchy, is the increased computational demand. Balancing the hierarchy depth with computational efficiency is an unexplored subject of investigation. Further, hierarchical GMM approaches often struggle with scalability and computational efficiency, particularly as the complexity of the model or the size of the data increases. 

The adaptive segmentation-wise initialization proposed in this paper divides the images into meaningful segments, and the parameters of the model are initialized based on the characteristics of these segments. This allows for a more localized and parallelized initialization approach, which can be advantageous in scenarios where hierarchical structures might become computationally burdensome.

\subsection{Block-based Initialization}
Block-based initialization methods \cite{tok_mse_2018,ozkan_steered_2023} divide an image into blocks and independently initialize parameters for each block. This can lead to more efficient computations and a potentially faster initialization phase. Verhack et al. \cite{verhack_universal_2016} introduced a block-based initialization method that leverages spatial activity analysis based on 2D-DCT. The method demonstrates good adaptability to spatial variations, providing fine-grained control over initialization. K-Means clustering algorithms have also been investigated for initialization and have shown very good results for block-based parameter initialization \cite{gul_big_2023,celeux_mixture_2007}. As a drawback, K-Means algorithms are highly sensitive to the initial random assignment of centroids. Although K-Means++ \cite{arthur_k-means_2007} addresses some issues related to random initialization, it often does not efficiently eliminate biases. As a consequence, the quality of initialization depends heavily on the nature of the data. Verhack et al. \cite{verhack_steered_2020} proposed to perform K-Means initialization on blocks instead of complete data to reduce the randomness. However, determining the correct number of clusters for K-Means can still be challenging for complex image block patterns, and an incorrect choice can lead to drastically suboptimal results.

A major advantage of block-wise initialization is the ability to parallelize computation. It is tempting to use optimized block SMoE models for the initialization of global full-image SMoE model optimization. Unfortunately, the above-mentioned ``global" nature of SMoE regression results in an undesirable impact of kernels in one block in remote areas of the image, because block SMoE models are usually not valid beyond their block-boundaries. In addition, block initialization leads to an excess of kernels, defeating the heavily desired sparsity functionality of the global SMoE model. 

While the segmentation of images into blocks is an excellent strategy to parallelize computation and reduce computational complexity and run-time, the method fails for SMoE models because real segmentation boundaries in images are not taken into account. The segmentation is arbitrary and has no relation to image content, particularly not with respect to edges in images. On the other hand, optimized SMoE models fragment images into segments with soft-gated region boundaries as shown in Fig \ref{fig1}. SMoE segments explicitly respect segment boundaries in images and place kernels to support these segment boundaries accordingly. The region-adaptive segmentation initialization strategy proposed and explored in this paper builds on the ``parallelization" capability of ``local" block approaches, by identifying segments in images that can be sufficiently explained with locally acting SMoE kernels. The hypothesis is, that segmentation boundaries with clear edges provide natural boundaries for kernels, avoiding to act beyond these segments. The now localized functioning of the kernels within pre-defined segments allows to initialize kernels per segment, rather than all kernels for all pixels.

\begin{figure*}[t]
\centering
\captionsetup{font=small}
\subfloat{\includegraphics[trim={0cm 0cm 0cm 0cm},clip,width=\linewidth]{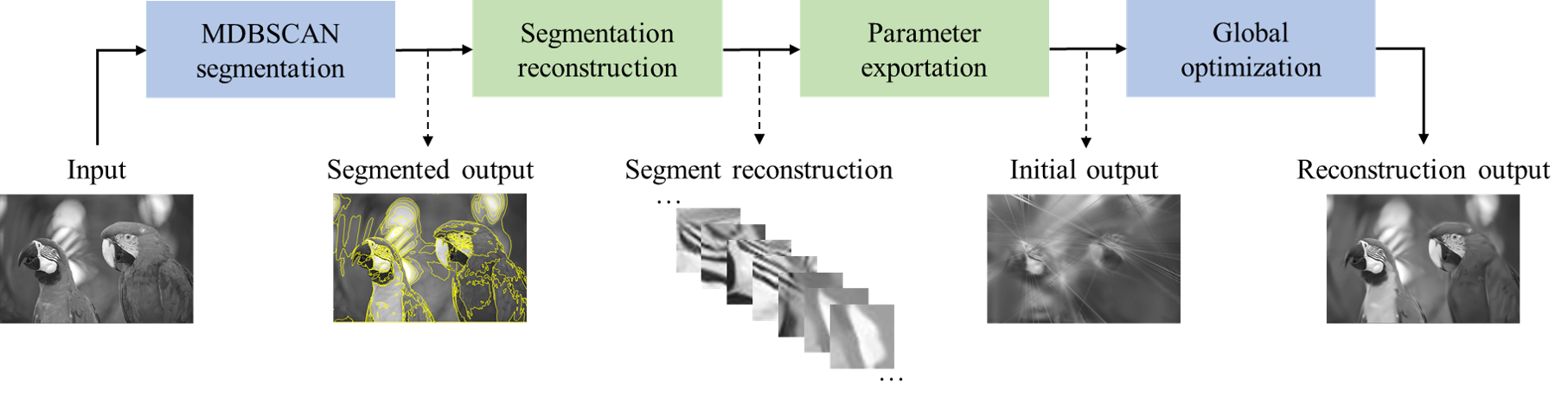}%
\label{barbara_psnr}}
\hfil
\vspace{-1em}
\caption{The overall pipeline of the proposed method. Two novel intermediate steps highlighted in green: Segmentation reconstruction and Parameter exportation are proposed to adaptively optimize the number of kernels as well as kernel parameters. The input data is segmented to dynamically adjust the kernels' number, location, and covariance in Segmentation reconstruction. This output then feeds into parameter exportation, where local parameters are aggregated, scaled, and prepared for global optimization, refining kernel parameters for efficient reconstruction.}
\vspace{-1em}
\label{fig2}
\end{figure*}

\subsection{Segmentation-based initialization}
Segmentation-based initialization methods have been a subject of interest and research \cite{legouhy_polaffini_2023,hu_refined_2023,li_segmentation-based_2023}, mostly not addressing the specific and delicate requirements of ``sum of weighted gates" regression with SMoE models. The idea of leveraging image segmentation techniques to guide the initialization of algorithms, such as image registration \cite{legouhy_polaffini_2023} or object recognition \cite{hu_refined_2023}, has been explored for several years. Segmentation-based methods can offer adaptive initialization, which is also at the heart of the proposed region-adaptive initialization method proposed in this paper. The adaptability can be crucial in scenarios where a uniform initialization may not capture the diverse characteristics of the data. Most segmentation-based initialization methods \cite{legouhy_polaffini_2023} leverage recent DL-based segmentation techniques to provide precise regional delineations. A deep network is trained to obtain semantic information. While it may be tempting to address initialization of SMoE model size and kernel parameters through ``learning", the complexity of images and the extremely large number of kernels makes such an approach unfeasible. 

In our initial work \cite{li_segmentation-based_2023}, Li et al. introduced the first attempt to use segmentation for initialization, specifically targeted for SMoE models. According to the characteristics of kernel regression methods, the output of the segment from a segmentation method should fulfill the following criteria: homogeneity and efficiency. The authors showed that the edge-based method is more suitable. Specifically, a simple MDBSCAN edge-based segmentation method outperforms other edge-based methods by solving the over-segmentation problem. 
The MDBSCAN algorithm was used to divide images into meaningful regions, followed by a random distribution of kernels within the segments. 
However, it relied on static segmentation, which overlooked inhomogeneity within and between segments, often resulting in suboptimal kernel placement and lacked flexibility in adapting to local image features. Consequently, enhanced adaptability to varying data distributions is required to address these limitations and improve overall performance.


\section{Adaptive Segmentation Initialization}

In this paper we will provide more insight and results for the segmentation initialization introduced in \cite{li_segmentation-based_2023}, and extend this basic MDBSCAN segmentation philosophy with the goal to arrive at drastically more accurate and efficient SMoE initialization. The MDBSCAN segmentation is specifically designed for kernel regression methods. It achieves homogeneity without fragmenting flat regions. While the initial methodology was rooted in a segmentation-centric paradigm, the proposed framework unfolds across two additional stages: 
\begin{itemize}
\item[a)] 
{\it Segmentation Reconstruction.} 
    The aim is to optimize the segmentation and segment initialization quality. The number of kernels and the kernel parameters are jointly optimized in a first optimization process. This avoids the limitations of the fixed-kernel approach in our previous work \cite{li_segmentation-based_2023}. By starting with regions that share similar characteristics, the local optimization process can find a more optimal solution more efficiently.
\item[b)] 
{\it Parameter Exportation.}
    This stage aims to explore localized insight gained from individual segments to global initialization of SMoE parameters in the full image domain. This stage is designed to serve as a pre-processing for the final subsequent global optimization process to arrive at the one large and final, global SMoE model.
\end{itemize}

The overall pipeline of the proposed method is depicted in Fig. \ref{fig2}. Starting with a MDBSCAN segmentation stage which is identical to our initial previous work in \cite{li_segmentation-based_2023}, images are divided into regions. The prime purpose is to arrive at regions with clearly defined boundaries, such as straddling edges or lines. The new {\it Segmentation Reconstruction} step processes each segment independently, resulting in segments with optimized SMoE parameters. To assemble these independent segments together for further global-wise optimization, the {\it Parameter Exportation} stage is proposed. This stage aggregates and processes all SMoE kernels and kernel parameters of all segments and provides a final SMoE initialization. Fig. \ref{fig2} illustrates individual segments processed at this stage. The final {\it Global Optimization} stage is identical to the work in \cite{li_segmentation-based_2023} and results in an optimal number of kernels and kernel parameters for regression.

In the following we address the motivation and algorithm details for the novel intermediate steps in Fig. \ref{fig2}.

\subsection{Segmentation Reconstruction}
The irregular nature of segment boundaries and the far-reaching impact of kernels beyond their assigned segments is addressed. 

\subsubsection{Cropping and Rescaling}
\begin{figure}[t]
\centering
\captionsetup{font=small}
\subfloat{\includegraphics[trim={0cm 0cm 0cm 0cm},clip,width=\linewidth]{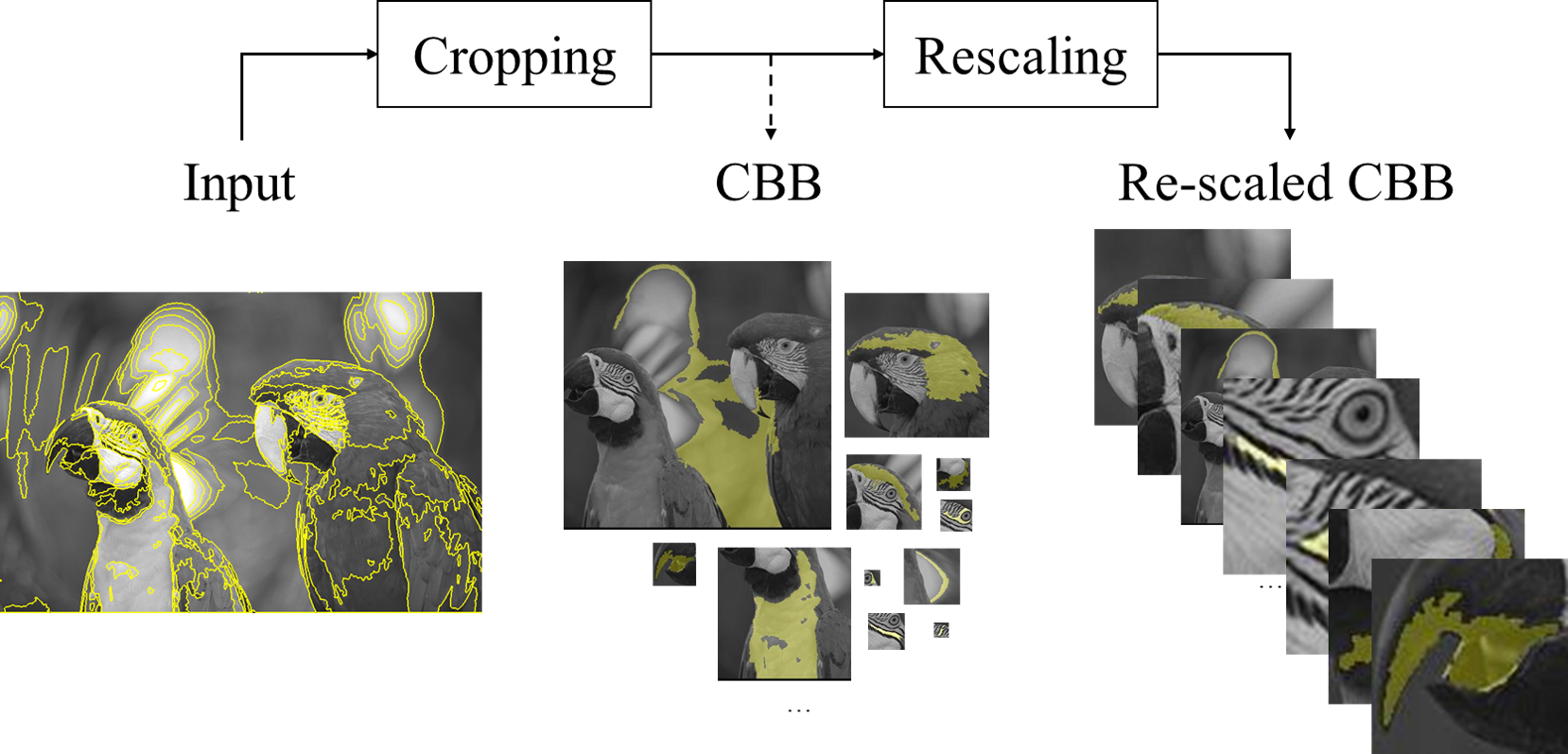}%
\label{barbara_psnr}}
\hfil
\vspace{-.5em}
\caption{Illustration of Cropping and Rescaling in Segmentation Reconstruction. The output of the cropping process results in a series of images containing the segmented region along with its surrounding pixels. The segmented region is highlighted yellow within each cropped image. Subsequently, these cropped images are resized to a standardized dimension to serve as input data for the local SMoE model.}
\vspace{-1em}
\label{fig3}
\end{figure}

This stage is defined to derive and process ``intermediate" segment representations to have similar Euclidean dimensions. The procedure of this stage is shown in Fig. \ref{fig3}. The motivation is twofold: normalizing the process and reducing complexity. This ``normalization" process allows the processing of all segments with similar processing parameters, such as GD step sizes and initial kernel bandwidths. This stabilizes the results of the subsequent adaptive-kernel optimization steps. In addition, we seek to down-scale segments to sizes with considerably less pixels compared to the original ones. In addition, parallel processing is now performed on well-defined and equal grids for all segments. Parallel processing on these normalized small sample bounding blocks allows to speed up computation considerably. 

The adaptive cropping process identifies the square box that can effectively contain all the pixels of a target segment. The size of the square box is determined by the maximum height and width of the segment, ensuring that the entire segment is encapsulated. Subsequently, a margin is added to all four sides of the square box to provide ample context. The addition of these corona pixels is required because the ``boundary" of a segment is not a well-defined line; rather, it is irregular and varies across different regions of the data. Additionally, the corona pixels serve as support for minimizing the ``global risc" - to avoid the external influence of kernels far outside of their segment domain. Each Corona Bounding Block (CBB) is re-scaled to a standardized size, identical to all CBBs. Since many original segments contain large number of pixels, this results in the massive down-sampling of many CBBs.

\subsubsection{Adaptive Kernel Adjustment}

In this step, a local Adaptive Kernel Adjustment is employed to understand how many kernels are required for each local CBB, and where to place the kernels. Adaptive number of kernels are allocated to each ``intermediate" segment depending on the complexity of the segment boundary and texture. This includes kernels placed at the outside of the segment boundaries. These ``supporting" kernels ensure that the boundaries within the segments are well preserved and that kernels in the segments have minimized global influence outside the segment. 

In essence, this step provides a local estimate of kernels that can be used for later parameter exportation and global optimization (depicted in Fig. \ref{fig2}), when kernels of local SMoE models of all CBBs are aggregated. At the core of this process is the attempt to control the quality of the ``global" reconstruction output of global optimization in later stages, by controlling the quality of reconstructed segments in this ``local" stage. For this purpose, for each CBB, a local SMoE model is optimized in an ``analysis-by-reconstruction" process. The optimal number of kernels and the kernel parameters in individual CBBs are unknown prior to optimization. These parameters, however, depend on the targeted quality of the reconstructed pixels in the CBB. Since individual CBBs contain segment boundaries and textures of varying complexity, different CBBs require different numbers of kernels to meet the same target qualities. The ``supporting" kernels are dropped from the SMoE models in later stages; when kernels of all segment models are combined to form the desired segment-adaptive initialization results for global optimization.


We employ a gradient descent optimization strategy to derive parameters for each local SMoE model by minimizing the CBB's mean square error (MSE) in conjunction with a kernel sparsification side condition. To meet a specific target quality, the optimization starts with a pre-defined initial number of kernels $L$. A dynamic kernel adjustment strategy is employed to ensure that a target PSNR is efficiently achieved even for CBBs with extremely complex segment boundaries. Algorithm \ref{alg:cap} details the initialization and iteration procedure used for the local Kernel Adjustment of each 
CBB. 

Notice, that in our specific implementation, the optimization and inversion of the kernel covariance matrices $\Sigma_j$ is avoided. Since the inverse of a symmetric matrix is again a symmetric matrix, we use a Cholesky decomposition $\Sigma_j^{-1}= B_j^{T} \cdot  B_j$ and estimate the parameters of the lower triangular matrix $B_j$ directly. This results in less parameters to optimize and generates significantly more stable iterations \cite{bochinski_regularized_2018}. Eq. (\ref{eq:4}) is then rewritten as
\begin{equation}
\label{eq:5}
w_j (x) = \frac{\pi_j \cdot \exp(-\frac{1}{2}(x-\mu_j)^T B_j^{T} B_j(x-\mu_j))}{{\textstyle\sum}_{i=1}^{K}\pi_i \cdot \exp(-\frac{1}{2}(x-\mu_i)^T B_i^{T} B_i (x-\mu_i))}
\end{equation}

During optimization, a crucial element is the utilization of $L1$ regularization loss \cite{bochinski_regularized_2018}, driving kernel sparsity by gradually reducing the weights $\pi_i$ of individual kernels until they approach zero. Our loss function $\mathcal{L}$ is then expressed for each CBB:

\begin{equation}
\label{eq:6}
\mathcal{L} = \frac{1}{card(S)}\sum_{x\in S} |y(x)-y_p(x)|^2 + \lambda \sum_{j=1}^{L}\pi_j,
\end{equation}
where $S$ are the pixels within the CBB, $card(S)$ is the number of pixels in the CBB, $y(x)$ is the intensity of original pixels, $y_p(x)$ is the SMoE regression, and $L$ is the initial number of kernels employed in the CBB respectively. $\lambda$ is an optimization hyper-parameter. 


\begin{algorithm}[t]
\vspace{5pt}
\caption{Adaptive Kernel Adjustment}\label{alg:cap}
\begin{algorithmic}
\REQUIRE Expert constant $M$, lower triangular matrix $B$, mean vector $U$, regression parameter $\pi$, terminal epoch $E$, target PSNR $Q$, initial number of kernels $L$, checkpoint $C$
\STATE $L \gets 10$
\STATE $M \gets RamdonVariable(L)$
\STATE $B \gets Ramdon2DMatrix(L)$
\STATE $U \gets RamdonVariable(L)$
\STATE $\pi \gets L^{-1}$
\STATE $currentLoss \gets 10^{6}$
\WHILE{iteration $i < E$}
    \STATE $y_i \gets SMoE(M,B,U,\pi)$
    \STATE $\mathcal{L} \gets MSE(y,y_i)+\lambda\sum\pi$
    \IF{$\mathcal{L} < currentLoss$}
        \STATE $Record(M,B,U,\pi)$
        \STATE $currentLoss \gets \mathcal{L}$
    \ENDIF
    \STATE $M,B,U,\pi \gets Adam(\Delta \mathcal{L})$
    \STATE $PSNR \gets 10\cdot\log_{10}(\mathcal{L}-\lambda\sum\pi)^{-1}$
    \IF{$PSNR > Q$}
        \RETURN $M,B,U,\pi$
    \ENDIF

    \IF{$i \mod C$ is 0}
        \STATE $M \gets Double(M)$
        \STATE $B \gets Double(B)$
        \STATE $U \gets Double(U)$
        \STATE $\pi \gets Double(\pi)$
        \STATE $L \gets Double(L)$
    \ENDIF
    
\ENDWHILE
\STATE $i \gets 0$
\WHILE{$i < L$}
    \IF{$\pi_i < 0$}
        \STATE $Remove(M_i,B_i,U_i,\pi_i)$
    \ENDIF
\ENDWHILE
\STATE $L \gets Size(M)$
\RETURN $M,B,U,\pi,L$
\end{algorithmic}
\label{algorithm1}
\end{algorithm}

Initially, the optimization is configured with a predefined number of kernels and then iterates over a specified number of cycles. After each iteration, the achieved PSNR is evaluated to determine the quality of the optimized result. If, after a pre-defined maximum number of iterations, the PSNR target is not reached, a dynamic adjustment is initiated. The number of kernels for the segment is doubled, effectively increasing the model's complexity. This adjustment is performed iteratively until the target PSNR is achieved or a predefined limit is reached. Finally, the kernels with weights $\pi_i$ below zero are eliminated.

By setting and striving for a target PSNR, and iteratively adjusting the model complexity, our approach ensures that the optimization process adapts to the diverse complexities inherent in different image segments, ultimately leading to more effective and efficient initialization of the SMoE model.

Even though it appears computationally cumbersome to optimize the many local SMoE models with GD at this stage, it is beneficial to recognize that few pixels are processed for each CBB. The CBBs in their intermediate format and size contain on average far less pixels compared to the original segments. This allows to derive local SMoE models for all CBBs using efficient, stable and fast parallel computation. 



\subsection{Parameter Exportation}

In a final step it is necessary to fuse all information from the local, small SMoE models into a global representation to start the global optimization of the desired one large SMoE model, which is responsible to explain the entire full original image. For this purpose, the kernels and kernel parameters from each  CBB are assembled to provide an efficient global initialization. The kernels from each CBB need to be placed into the original full size pixel grid. It is, however, neither possible nor desired to use the ``local" SMoE model kernels and parameters ``as is" to assemble a large set of all local kernels for initializing the one ``global" SMoE model optimization. 

The primary purpose of the kernel fusion in the Parameter Exportation stage is to construct a robust bridge between the local knowledge represented by down-sampled segment-domain parameters and the global context of the entire data. The exportation process seeks to achieve consistency and coherence in parameter representation across the entire data. It ensures that the knowledge gained from each down-sampled segment contributes meaningfully to the overall understanding without creating disjointed or conflicting representations. We notice, that:
\begin{itemize}
\item[-] 
the local SMoE models live in a heavily down-sampled CBB domain, and thus their kernel parameters need to be adjusted to the size of the respective individual segments.
\item[-] 
a large abundant number of kernels exist in the local SMoE models that are redundant for global initialization. Since it is a prime focus to arrive at a sparse global model, these redundant kernels need to be removed.
\end{itemize}

\subsubsection{Redundant Kernels}

After optimization of the individual local SMoE models, the kernels that lay outside of the intermediate segment boundaries are no longer required for a global model. The reason is that the kernels of adjacent segments act as natural supporting kernels for the segment and, after the kernels of a down-sampled CBB are up-sampled, to fit the original segment size and position. Therefore, the kernels lying outside of the segment boundaries are redundant after kernel fusion and are removed as they no longer contribute to the representation of the segment.

\subsubsection{Steering Kernel Up-Sampling}

The position and parameters of the optimized kernels that lay within a CBB segment need to be up-scaled to fit the original segment size. It is of vital importance, that the topological structure of the segment boundaries and within-segment textures are preserved. 
As the parameters move from the low-resolution segment domain to the entire full-resolution image domain, an adaptive transformation is necessary. Scaling a 2D Gaussian involves multiplying the original covariance matrix $\Sigma$ by a scaling factor $s$, $\Sigma_s = s \cdot \Sigma$. Alternatively, the scaled lower diagonal matrix is  $B_s = 1/\sqrt{s} \cdot B$.

\begin{table}[t]
\captionsetup{font=small,justification=centering, labelsep=newline,textfont=sc}
\caption{Average Reconstruction Quality over segments}
\begin{tabular}{>{\centering}p{.08\textwidth}|>{\centering}p{.07\textwidth}>{\centering}p{.07\textwidth}>{\centering}p{.07\textwidth}>{\centering\arraybackslash}p{.07\textwidth}}
\hline
 \#Epoch & 100 & 200 & 300 & 400 \\
\hline
PSNR [dB]	&18.37&	21.33	&23.13&	23.70 \\
Time (s) 	&6.99	&19.83	&44.32	&92.14\\
\hline
\end{tabular}
\label{table1}
\end{table}

\section{Experimental Setting}

We investigated the algorithmic framework using gray-scale versions of test images ``Barbara" (720x576 pixels), ``Flower" (768x512 pixels), ``Parrot" (768x512 pixels) and ``Peppers" (512x512 pixels).

\subsection{Parameter Settings in Segmentation Reconstruction}

\begin{figure}[t]
\captionsetup{font=small}
\centerline{\includegraphics[trim={0cm 0cm 2cm 2.1cm},clip,width=1.05\linewidth]{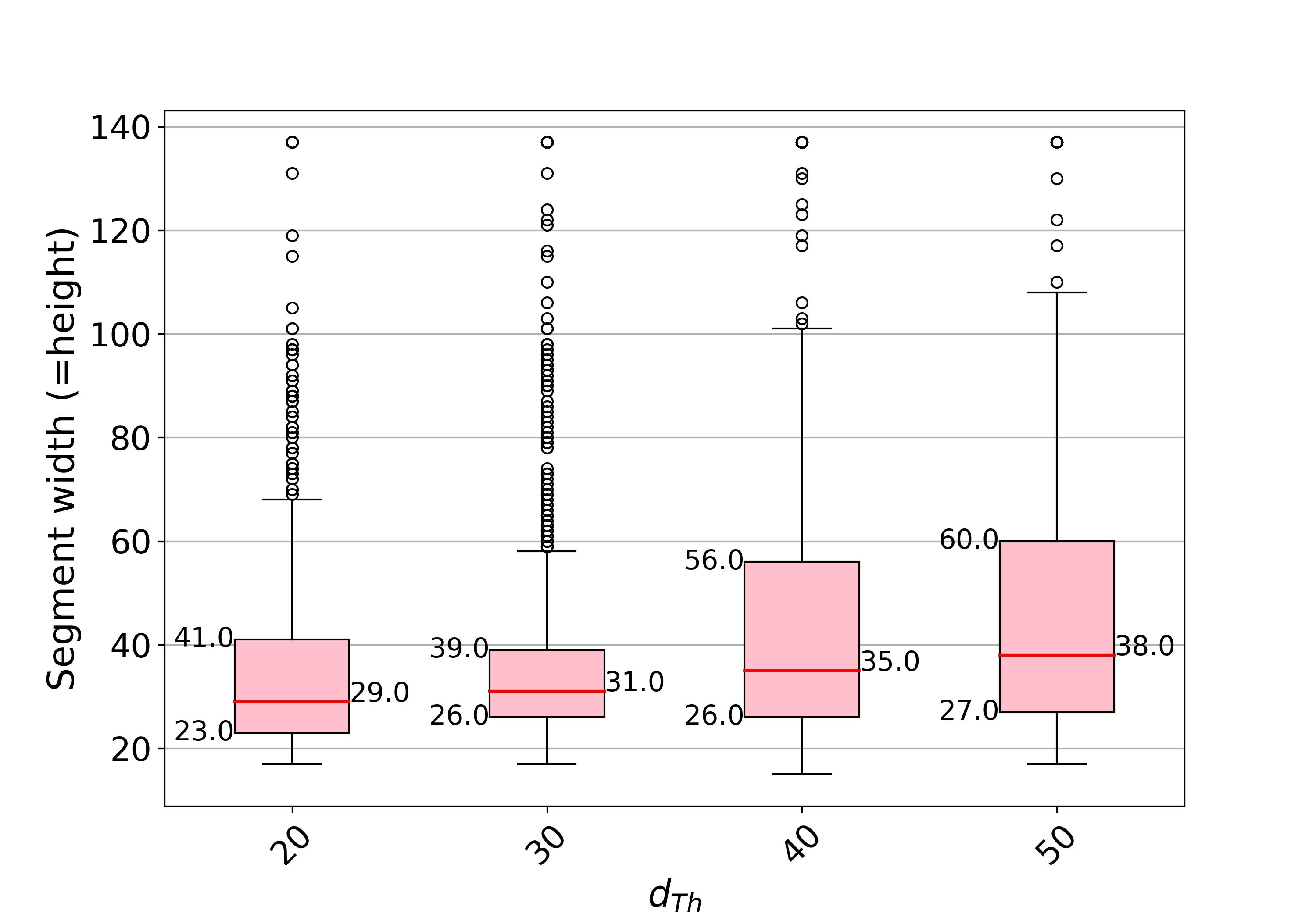}}
\vspace{-0.5em}
\caption{Pixel intensity difference threshold $d_{Th}$ versus the width (=height) of the cropped segments. The generated size of the segments (measured as pixel width=height of a bounding box) is in the range between 20 and 40 for the different thresholds $d_{Th}$.}
\vspace{-1em}
\label{fig4}
\end{figure}

The MDBSCAN clustering algorithm is used in this paper for image superpixel segmentation. One tunable parameter is the pixel difference threshold $d_{Th}$. A higher threshold enforces not only flat areas but also regions with intricate textures. Usually only a few large segments are generated per image. Conversely, a lower threshold may yield many excessively fragmented segments with minimal pixel differences. The box plot in Fig. \ref{fig4} illustrates the impact of four difference thresholds $d_{Th}$ on the generated size of the segments (measured as pixel width=height of a bounding box). We observe that the average segment width or height is in the range between 20 to 40 pixels for thresholds between 20-40. Even a larger difference threshold of 50 results in an average width smaller 40 pixels. For our purposes, it is desirable to arrive at segments that contain mainly flat textures. Thus, the SMoE kernels in the segments mainly focus on reconstruction of the segment boundaries, which are the relevant edges and lines in the images. On the other hand it is beneficial to arrive at segments that are not too small. 

From the experiments, a segment size of 32x32 appeared as a reasonable compromise and was used for further investigation as a bounding block size (thus CBB size of 32x32 pixels). Thresholds between 40-50 were used for further experimentation to limit the number of segments per image. For the given thresholds, the size of most segments drastically exceeded the bounding block size 32x32, as can be observed from Fig. \ref{fig4}. Thus, the down-sampling of most segments resulted in the massive reduction of samples desired for efficient parallel processing of the data.

Table \ref{table1} provides insight into the achieved reconstruction quality after different numbers of epochs and the corresponding run-time in the Segmentation Reconstruction phase on the 32x32 bounding blocks. PSNR results are averaged over all test segments, and the run-time is the total measured for all segments. A total number of 1666 segments were processed for the four test images in this experiment. From the observation of this experimental result, a termination criterion of 300 epochs for each segment was selected as a practical compromise, ensuring that the computational resources are utilized efficiently without sacrificing notable gains in PSNR.

Another parameter that needs to be specified is the ``corona" margin. The purpose of the margin is primarily to mitigate long-distance effects while ensuring that surrounding pixels are included without significantly increasing complexity. Therefore, we have opted for a conservative margin size to balance these considerations. In this work, we have found that a margin of 5 pixels suffices for our purposes. However, further analysis of this parameter setting could be conducted for more comprehensive insights.

\begin{table}[t]
\captionsetup{font=small,justification=centering, labelsep=newline,textfont=sc}
\caption{Initialization setting}
\begin{tabular}{>{\centering}m{.065\textwidth}|>{\centering}m{.198\textwidth}>{\centering\arraybackslash}m{.158\textwidth}}
\hline
  Method & {Kernel Type} & {Kernel Parameter Output} \\
\hline
Grid & Radial, all same bandwidth & $\mu_{j}$, Diagonal Matrix \\
K-Means & Radial, all same bandwidth & $\mu_{j}$, Diagonal Matrix \\
S-SMoE  & Radial, all same bandwidth & $\mu_{j}$, Diagonal Matrix \\
AS-SMoE & Steering, all individual steering & $\mu_{j}$, $\Sigma_{j}$ \\
\hline
\end{tabular}

\vspace{-1em}
\label{table2}
\end{table}

\begin{table*}[tb]
\vspace{10pt}
\captionsetup{font=small,justification=centering, labelsep=newline,textfont=sc}
\caption{Reconstruction Quality with different number of optimal kernels $L_{Opt}$}
\begin{tabular}{>{\centering}m{.04\textwidth}|>{\centering}m{.02\textwidth}>{\centering}m{.034\textwidth}>{\centering}m{.034\textwidth}>{\centering}m{.034\textwidth}|>{\centering}m{.02\textwidth}>{\centering}m{.034\textwidth}>{\centering}m{.034\textwidth}>{\centering}m{.034\textwidth}|>{\centering}m{.015\textwidth}>{\centering}m{.02\textwidth}>{\centering}m{.034\textwidth}>{\centering}m{.034\textwidth}>{\centering}m{.034\textwidth}|>{\centering}m{.015\textwidth}>{\centering}m{.02\textwidth}>{\centering}m{.034\textwidth}>{\centering}m{.034\textwidth}>{\centering\arraybackslash}m{.034\textwidth}}
\hline
  & \multicolumn{4}{c|}{Grid} & \multicolumn{4}{c|}{K-Means} & \multicolumn{5}{c|}{S-SMoE} & \multicolumn{5}{c}{AS-SMoE} \\
\hline
Image &  $L_{Opt}$ & PSNR & SSIM & LPIPS &  $L_{Opt}$ & PSNR & SSIM & LPIPS & $d_{Th}$& $L_{Opt}$ & PSNR & SSIM & LPIPS   &  $d_{Th}$& $L_{Opt}$&  PSNR  & SSIM & LPIPS\\
\hline
\multirow{4}{*}{barbara}&\multirow{2}{*}{3279} &\multirow{2}{*}{{23.12}} & \multirow{2}{*}{0.62}& \multirow{2}{*}{0.51} & 
\multirow{2}{*}{3186} &\multirow{2}{*}{{22.71}} 
& \multirow{2}{*}{0.61}& \multirow{2}{*}{0.59} & 
                         \multirow{2}{*}{40} & 2824 &{23.10} 
                         & 0.63& 0.53 & \multirow{2}{*}{40} & 2922 &\textbf{24.66} 
                         &0.71 & 0.38\\
                                      &&&&&&&&&& 1379 &{23.41} 
                                      & 0.63& 0.52 &                     & 1653 &\textbf{24.81} 
                                      &0.70 & 0.38\\
                        &\multirow{2}{*}{1926} &\multirow{2}{*}{{22.87}} 
                        & \multirow{2}{*}{0.61}& \multirow{2}{*}{0.53}  & \multirow{2}{*}{1822} &\multirow{2}{*}{{22.48}} 
                        & \multirow{2}{*}{0.58}& \multirow{2}{*}{0.61}  & 
                         \multirow{2}{*}{50} & 2516 &{23.42} 
                         & 0.64& 0.51 & \multirow{2}{*}{50} & 2659 &{24.09} 
                         &0.68 & 0.41\\
                                       &&&&&&&&&& 1714 &{23.52} 
                                       & 0.63& 0.52 &                     & 1939 &{24.72} 
                                       &0.71 & 0.38\\
\hline
\multirow{4}{*}{flower} &\multirow{2}{*}{1619} &\multirow{2}{*}{{24.49}}	
&\multirow{2}{*}{0.71}& \multirow{2}{*}{0.45}  & \multirow{2}{*}{1956} &\multirow{2}{*}{{26.10}}	
&\multirow{2}{*}{0.76}& \multirow{2}{*}{0.40}  & 
                         \multirow{2}{*}{40} & 2031  &{27.71}	
                         &	0.80& 0.35 & \multirow{2}{*}{40} & 2094 &\textbf{29.09} 
                         &0.84 & 0.28\\
                                       &&&&&&&&&& 1060  &{26.62} 
                                       & 0.76& 0.41 &                      & 1178 &\textbf{28.54} 
                                       &0.82 & 0.30\\
                        &\multirow{2}{*}{1101} &\multirow{2}{*}{{24.11}}	
                        &\multirow{2}{*}{0.69} &\multirow{2}{*}{0.48} &\multirow{2}{*}{1334} &\multirow{2}{*}{{25.19}}	
                        &\multirow{2}{*}{0.72} &\multirow{2}{*}{0.45} & 
                         \multirow{2}{*}{50} & 1414  &{26.60} 
                         &	0.76& 0.41 & \multirow{2}{*}{50} & 1591 &{29.14} 
                         &0.84 & 0.29\\
                                       &&&&&&&&&& 767   &{25.44} 
                                       & 0.71& 0.48 &                      & 960 &{28.08} 
                                       &0.81 & 0.32\\
\hline

\multirow{4}{*}{parrot} & \multirow{2}{*}{1322} &\multirow{2}{*}{{27.23}}	
& \multirow{2}{*}{0.82}&	\multirow{2}{*}{0.55} &\multirow{2}{*}{1468} &\multirow{2}{*}{{28.31}}	
& \multirow{2}{*}{0.83}&	\multirow{2}{*}{0.52} & 
                         \multirow{2}{*}{40} & 1134  &{29.39}	
                         &	0.84& 0.49 & \multirow{2}{*}{40} & 1203 &\textbf{32.22} 
                         &0.86 & 0.42\\
                                       &&&&&&&&&& 916  &{29.13} 
                                       & 0.83& 0.51 &                     &  896 &\textbf{31.39} 
                                       &0.85 & 0.44\\
                        &\multirow{2}{*}{826} &\multirow{2}{*}{{26.88}}	
                        &\multirow{2}{*}{0.81}&\multirow{2}{*}{0.56} & \multirow{2}{*}{893} &\multirow{2}{*}{{28.18}}	
                        &\multirow{2}{*}{0.82}&\multirow{2}{*}{0.54} & 
                         \multirow{2}{*}{50} &752 &{28.45} 
                         &0.82 &0.53 &\multirow{2}{*}{50} &861 &{31.55} 
                         &0.85 & 0.45\\
                                       &&&&&&&&&& 645  &{28.06} 
                                       & 0.82& 0.55 &                   &626 &{30.47} 
                                       &0.84 & 0.47\\
\hline

\multirow{4}{*}{peppers}&\multirow{2}{*}{1222} &\multirow{2}{*}{{26.47}} 
&\multirow{2}{*}{0.74}& \multirow{2}{*}{0.55}  & \multirow{2}{*}{1415} &\multirow{2}{*}{{27.90}} 
&\multirow{2}{*}{0.78}& \multirow{2}{*}{0.51}  & 
                         \multirow{2}{*}{40} & 1440 &{29.46} 
                         &0.79 &0.48 & \multirow{2}{*}{40} & 1346 &\textbf{31.66} 
                         &0.82 & 0.42\\
                                       &&&&&&&&&& 864  &{28.30} 
                                       &0.78 &0.50 &                      &  806 &\textbf{30.17} 
                                       &0.80 & 0.44\\

                        &\multirow{2}{*}{776}    &\multirow{2}{*}{{25.73}}	
                        &\multirow{2}{*}{0.73}&\multirow{2}{*}{0.56}  &\multirow{2}{*}{850}    &\multirow{2}{*}{{27.66}}	
                        &\multirow{2}{*}{0.77}&\multirow{2}{*}{0.53}  & 
                         \multirow{2}{*}{50} &  959  &{28.84} 
                         &0.78& 0.50 & \multirow{2}{*}{50} & 992 &{30.39} 
                         &0.80 & 0.44\\
                                       &&&&&&&&&& 588  &{27.93}   
                                       &0.76 &0.53 &                      &  631 &{29.10} 
                                       &0.78 & 0.47\\
\hline
\end{tabular}

\vspace{1em}
\label{table3}
\end{table*}

\subsection{Evaluation criteria}

In our study both image reconstruction quality as well as computational aspects were evaluated. In addition to established SSIM and PSNR mesures, the PSNR was calculated using a Vision Transformer Attention Map. The LPIPS metric was also evaluated on the test set.

The Learned Perceptual Image Patch Similarity (LPIPS) \cite{zhang_unreasonable_2018} has recently received much attention as an objective measure. Apparently more tuned to human visual perception than e.g. the PSNR, this metric captures suttled nuances in image quality. A pre-trained Vision Transformer \cite{dosovitskiy_image_2021} used to derive an attention map highlights regions of interest from a human perspective. By applying attention weighting to the PSNR computation, this metric provides an evaluation focused on regions of interest. The attention weighting is determined by scaling the attention map between 0.9 and 1.

We compare the novel adaptive segmentation-based SMoE initialization strategy AS-SMoE with previous segmentation-based SMoE initialization work (S-SMoE) in \cite{li_segmentation-based_2023}, which already improved significantly over the state-of-the-art.  The method in \cite{li_segmentation-based_2023} did not derive the adaptive number of kernels for each segment, and relied on fixed kernel distributions. In addition, the naive regular grid (Grid)\cite{bochinski_regularized_2018} and well-known and previously well-established SMoE block-based K-Means initialization \cite{verhack_steered_2020} was used as benchmark. The details of the initialization setting of three methods are shown in Table \ref{table2}.

\begin{figure*}[th]
\centering
\captionsetup{font=small}
\begin{tabularx}{\textwidth} { 
   >{\centering\arraybackslash}p{0.11\linewidth}
   >{\centering\arraybackslash}p{0.11\linewidth}
   >{\centering\arraybackslash}p{0.16\linewidth}
   >{\centering\arraybackslash}p{0.16\linewidth} 
   >{\centering\arraybackslash}p{0.16\linewidth}
   >{\centering\arraybackslash}p{0.16\linewidth}}
    \includegraphics[trim={0cm 0cm 0cm 0cm},clip,width=1.1\linewidth]{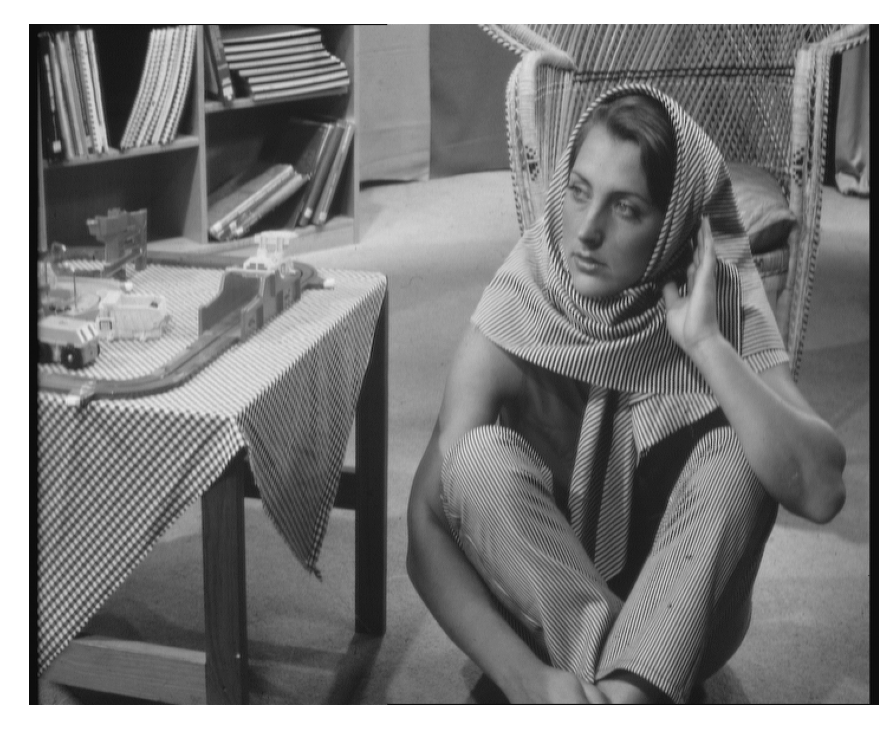} & 
    \includegraphics[trim={1.5cm 0cm 1.5cm 0cm},clip,width=1.1\linewidth]{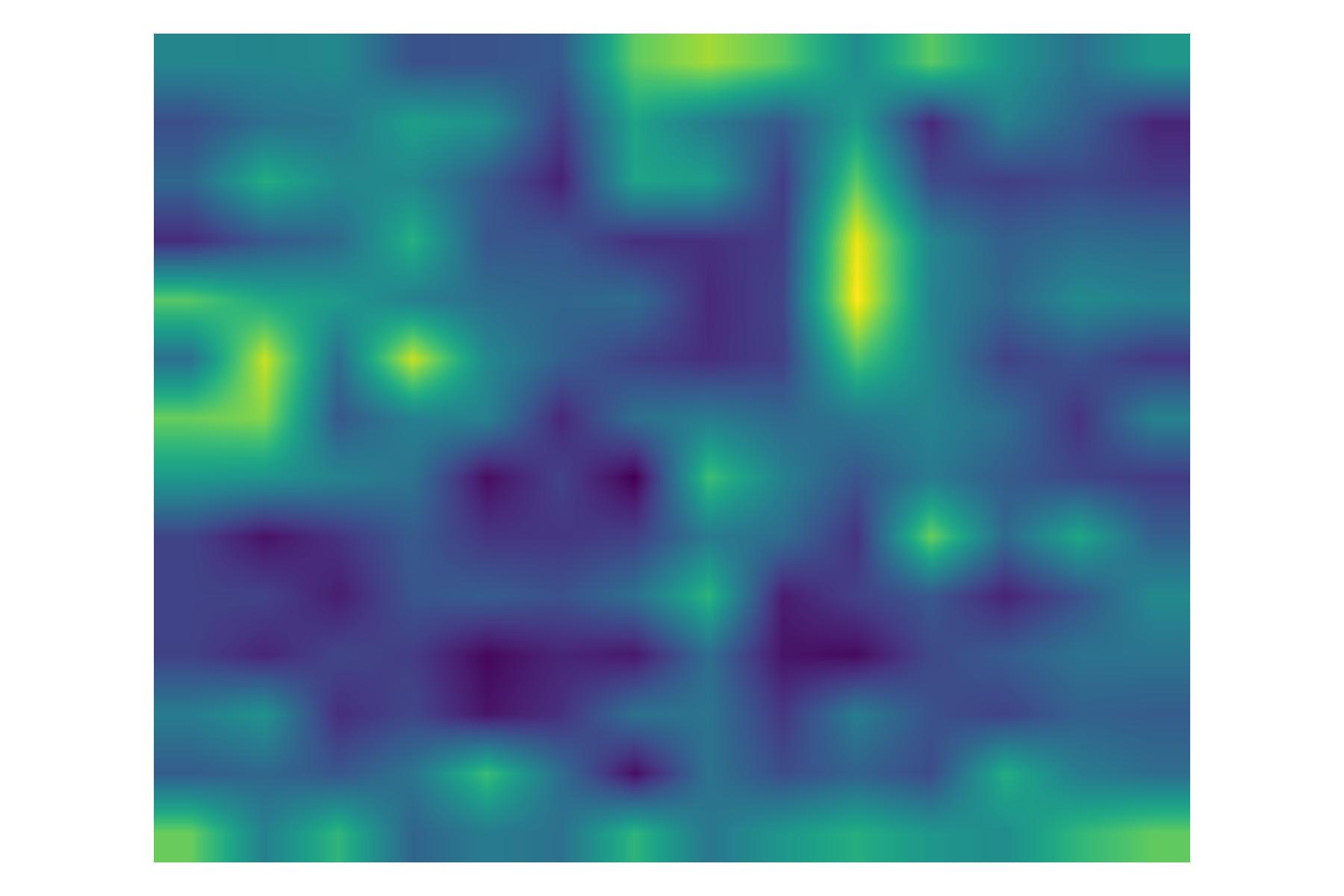} & 
    \multirow{2}{*}[.7in]{\includegraphics[trim={0cm 0.5cm 0cm 0.5cm},clip,width=1.18\linewidth]{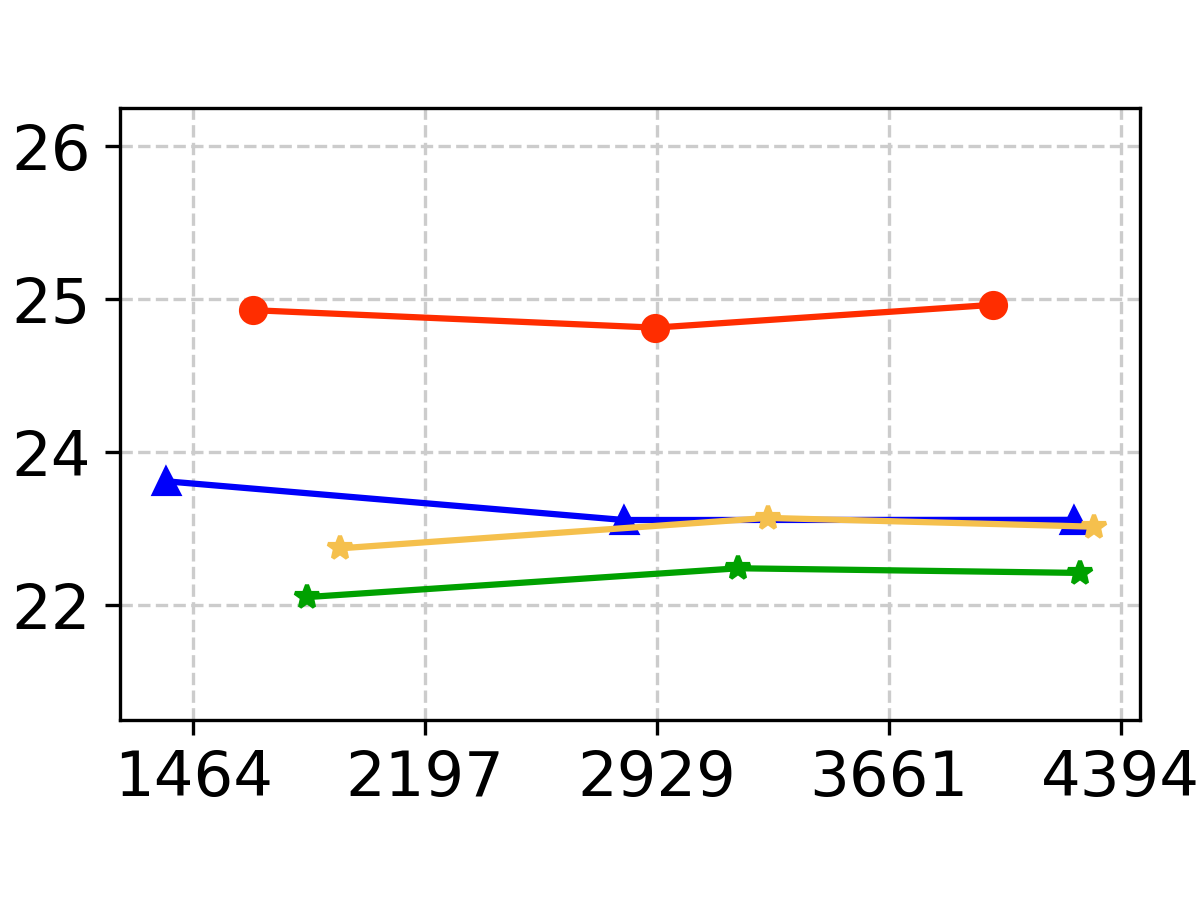}} & 
    \multirow{2}{*}[.7in]{\includegraphics[trim={0cm 0.5cm 0cm 0.5cm},clip,width=1.18\linewidth]{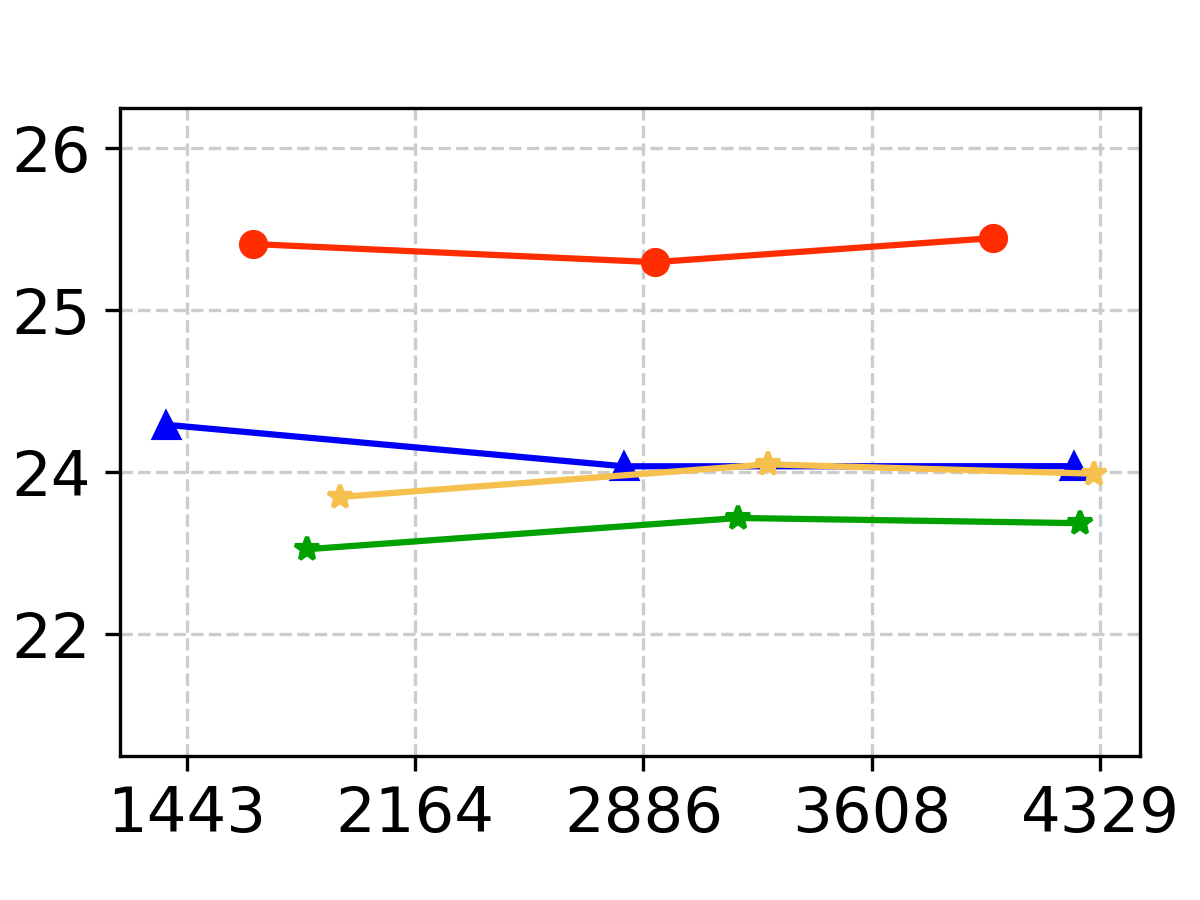}} &
    \multirow{2}{*}[.7in]{\includegraphics[trim={0cm 0.5cm 0cm 0.5cm},clip,width=1.18\linewidth]{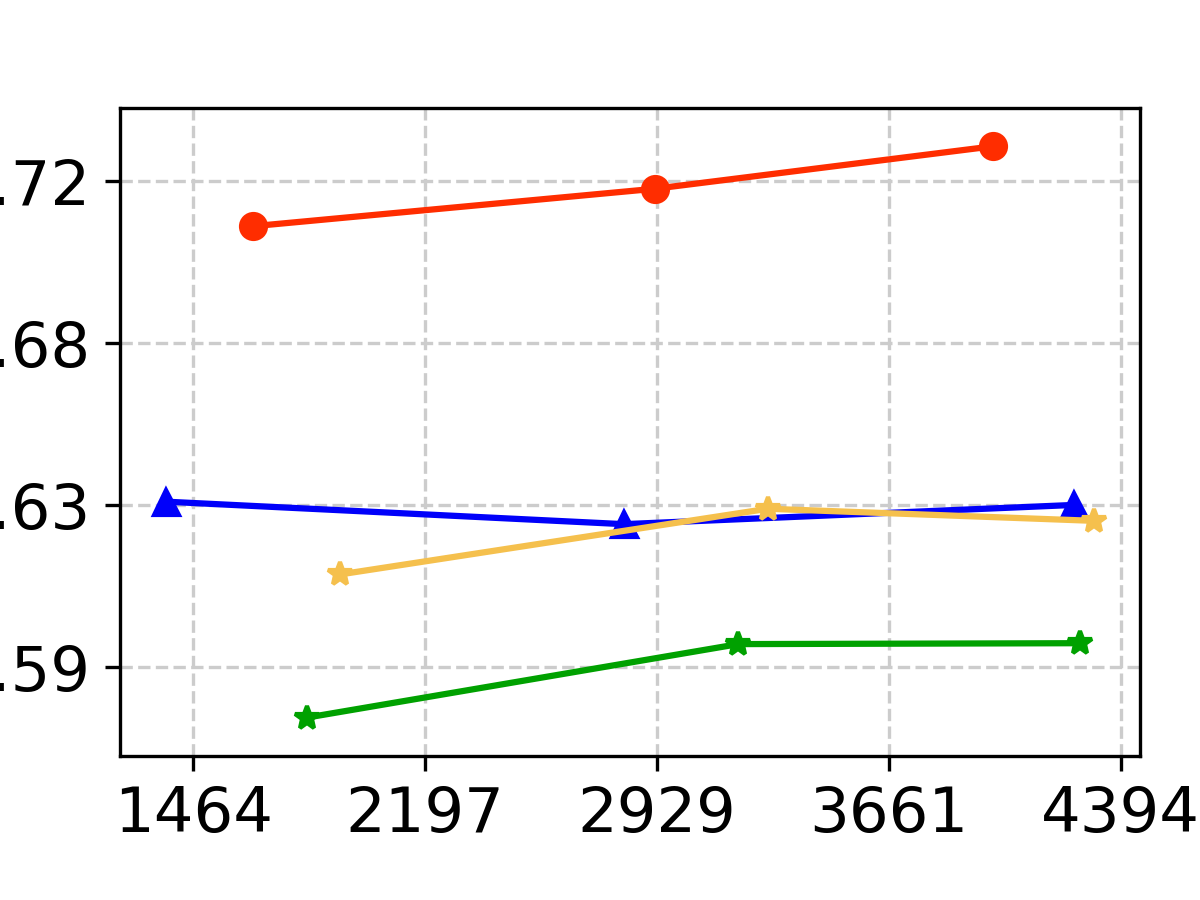}} &
    \multirow{2}{*}[.7in]{\includegraphics[trim={0cm 0.5cm 0cm 0.5cm},clip,width=1.18\linewidth]{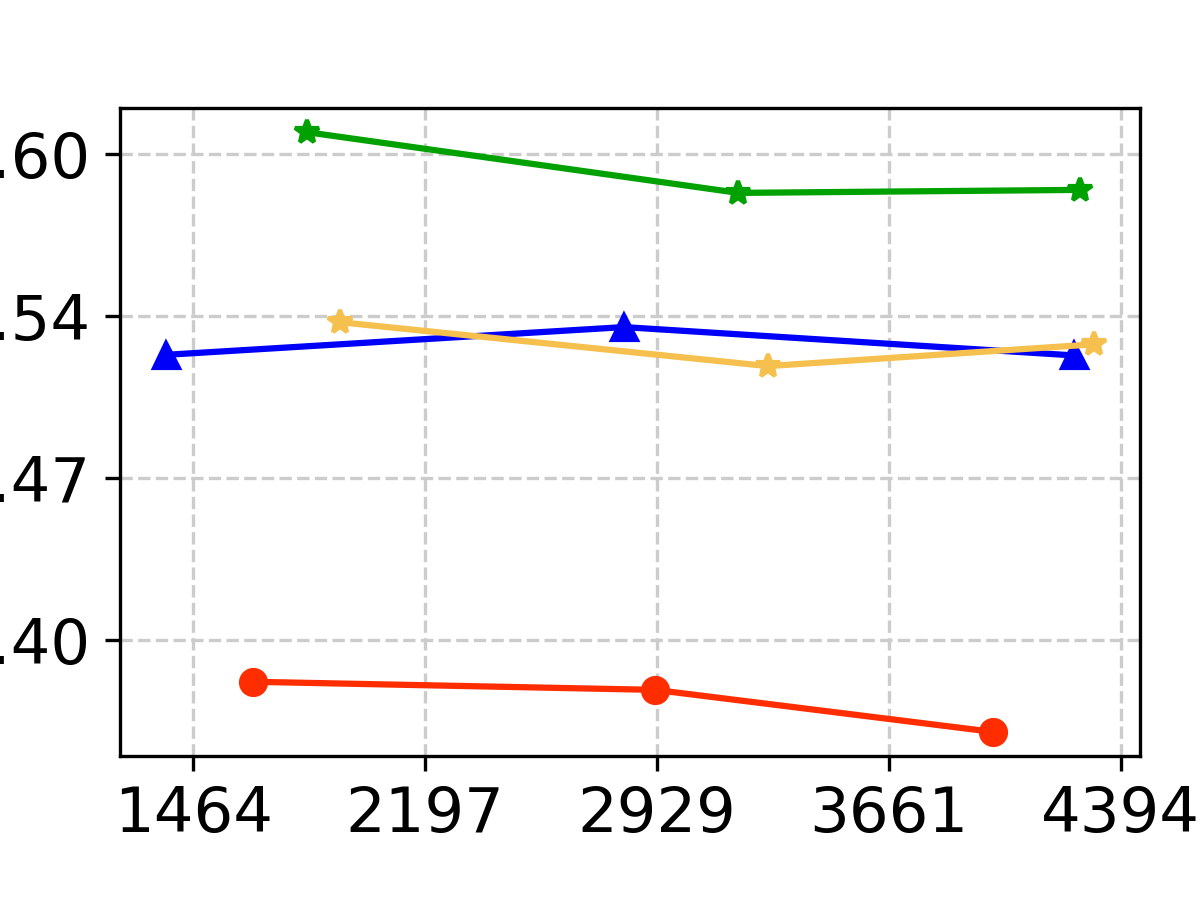}} \\
    \small{barbara} &  & & \\
    
    \\
    \includegraphics[trim={0cm 0cm 0cm 0cm},clip,width=1.1\linewidth]{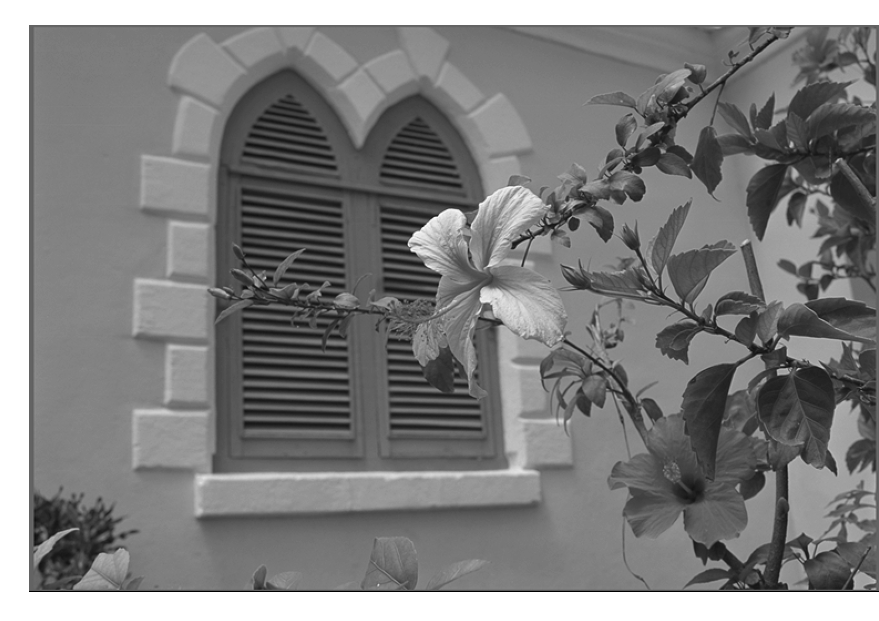} & 
     \includegraphics[trim={0.5cm 0cm 0.5cm 0cm},clip,width=1.1\linewidth]{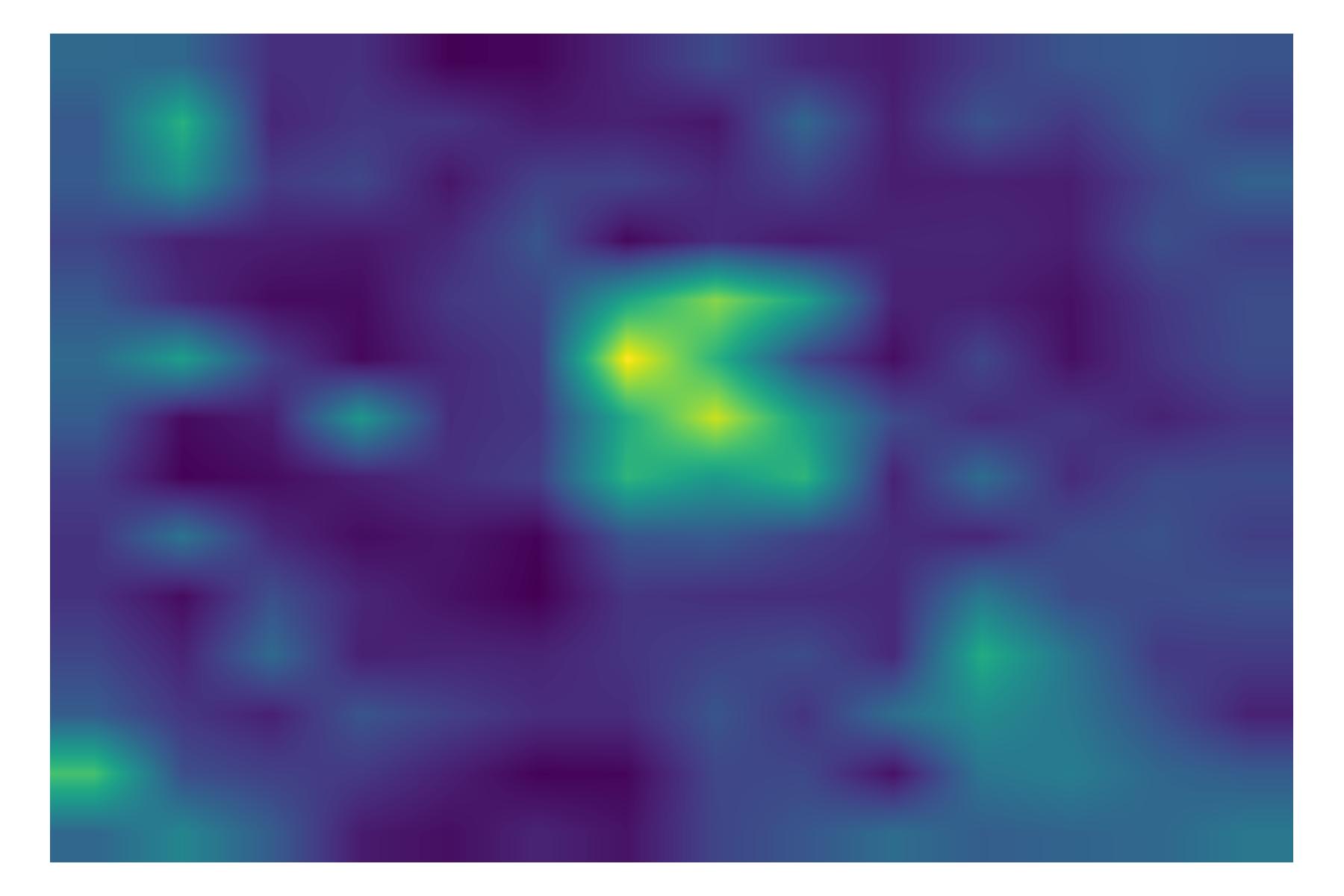} & 
    \multirow{2}{*}[.65in]{\includegraphics[trim={0cm 0.5cm 0cm 0.5cm},clip,width=1.18\linewidth]{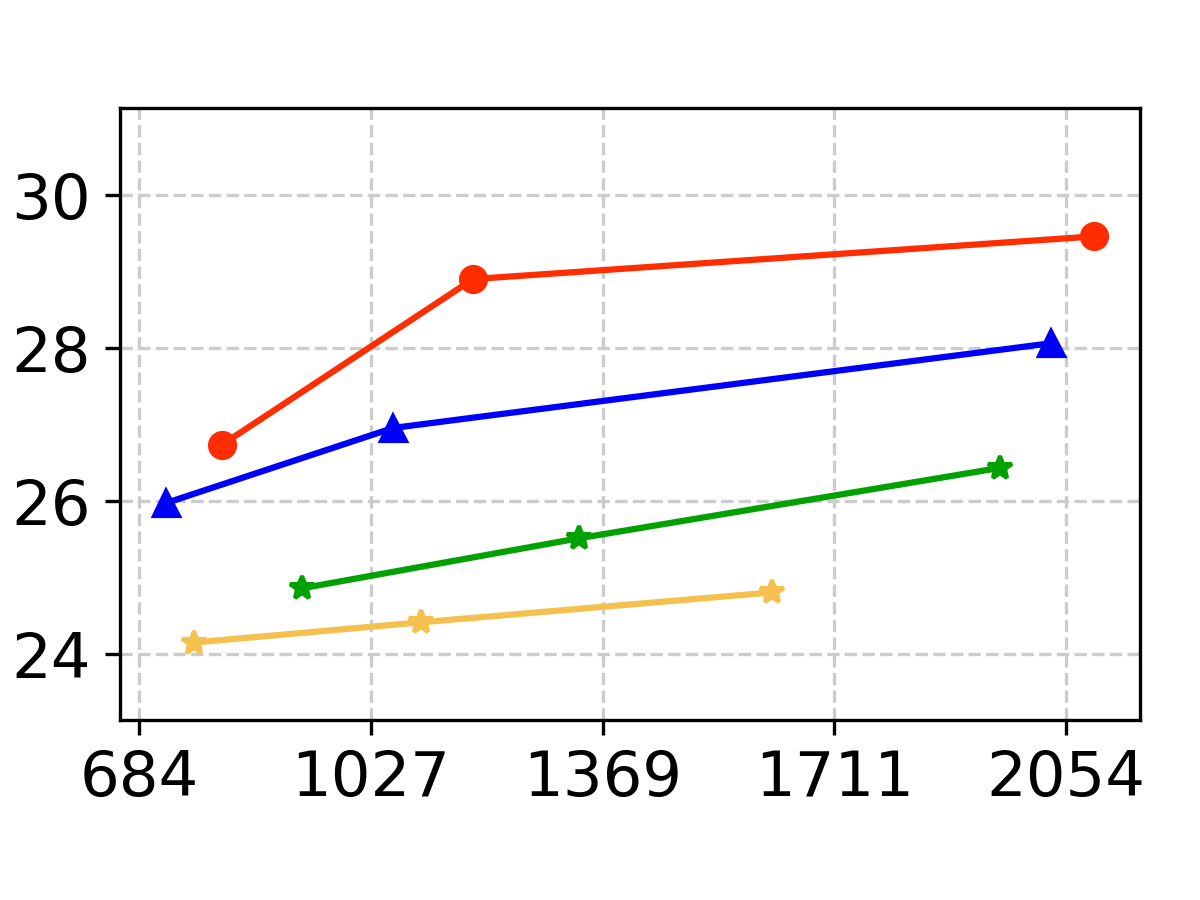}} & 
    \multirow{2}{*}[.65in]{\includegraphics[trim={0cm 0.5cm 0cm 0.5cm},clip,width=1.18\linewidth]{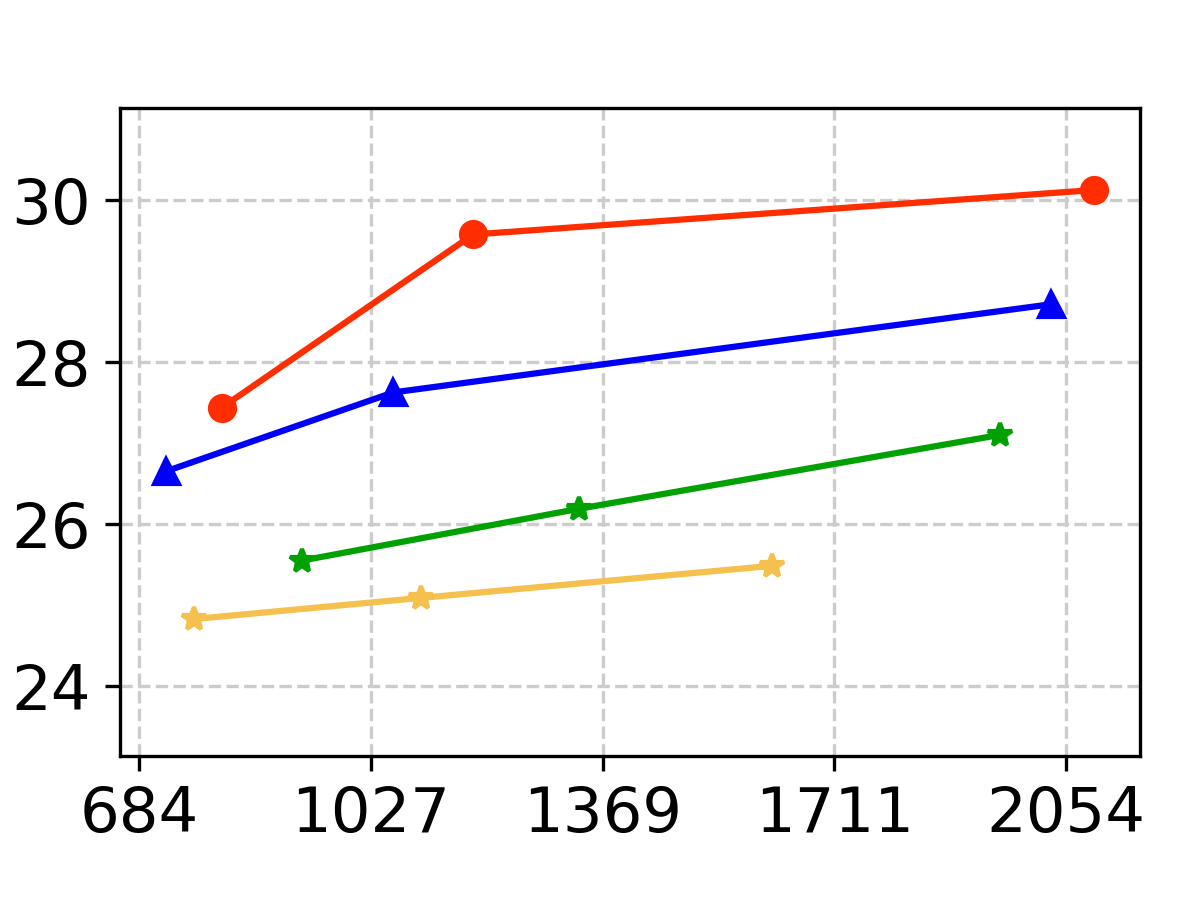}} &
    \multirow{2}{*}[.65in]{\includegraphics[trim={0cm 0.5cm 0cm 0.5cm},clip,width=1.18\linewidth]{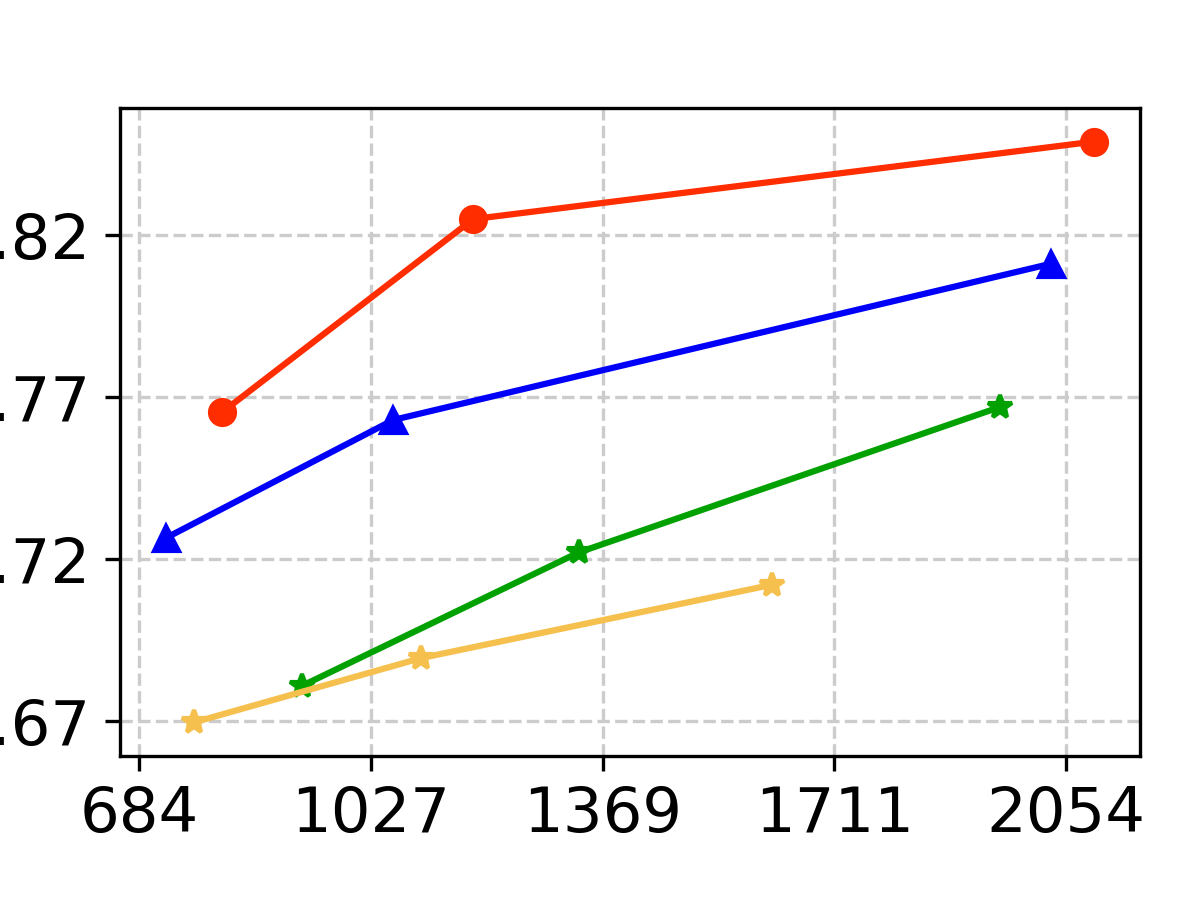}} &
    \multirow{2}{*}[.65in]{\includegraphics[trim={0cm 0.5cm 0cm 0.5cm},clip,width=1.18\linewidth]{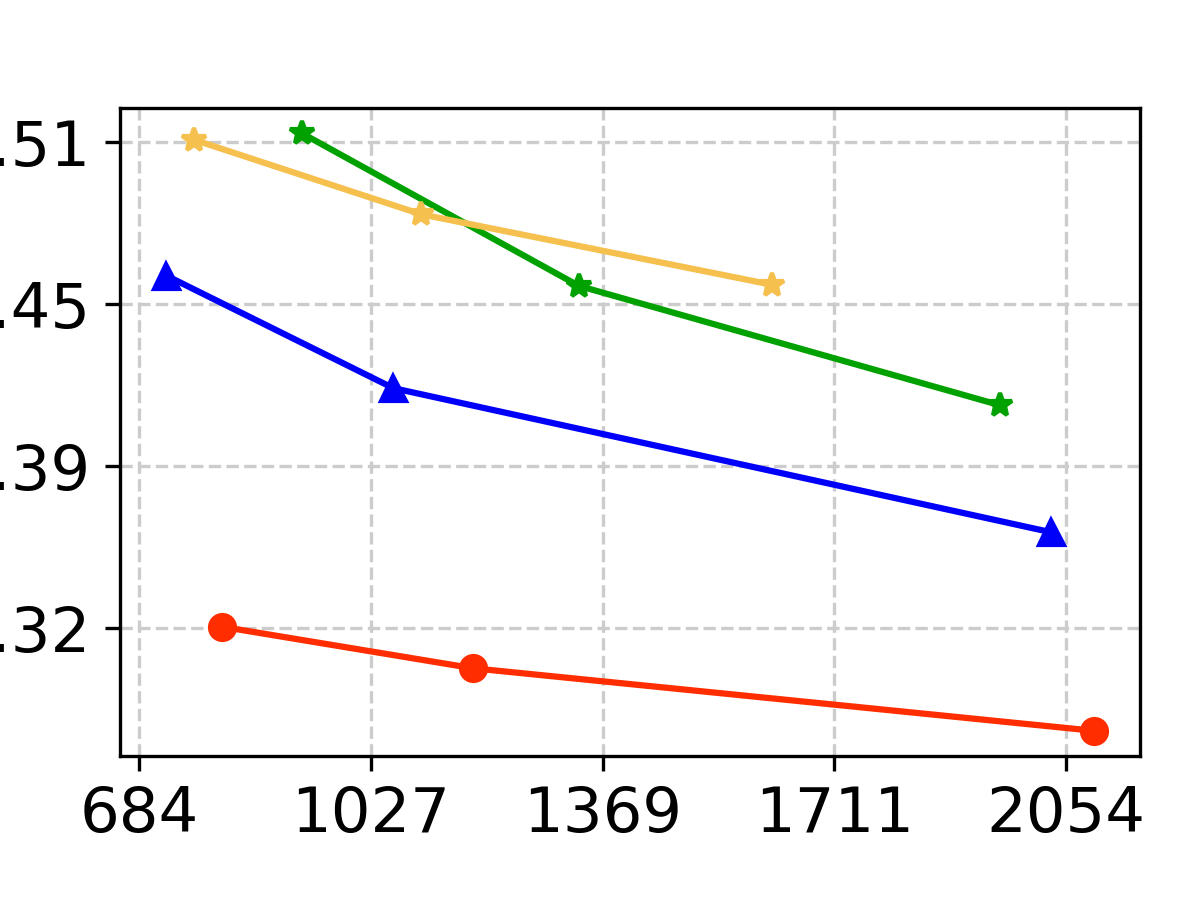}} \\
    \small{flower} &  & & \\
    \\
    \includegraphics[trim={0cm 0cm 0cm 0cm},clip,width=1.1\linewidth]{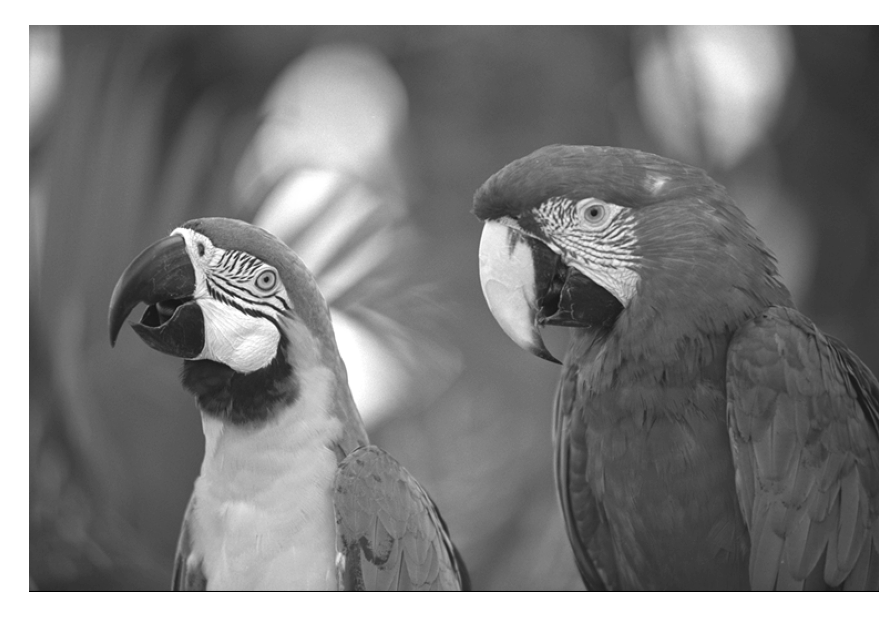} & 
     \includegraphics[trim={0.5cm 0cm 0.5cm 0cm},clip,width=1.1\linewidth]{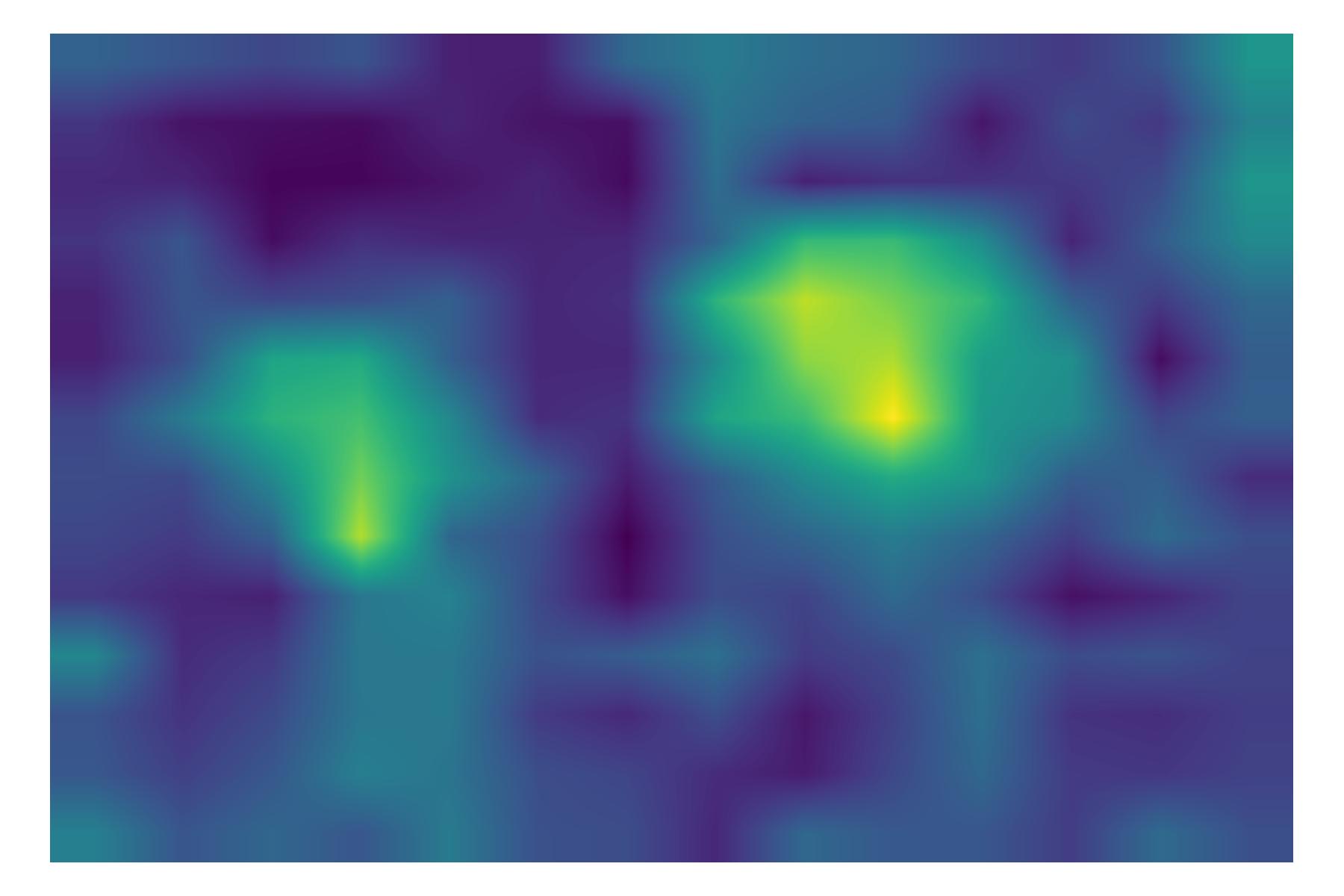} & 
    \multirow{2}{*}[.6in]{\includegraphics[trim={0cm 0.5cm 0cm 0.5cm},clip,width=1.18\linewidth]{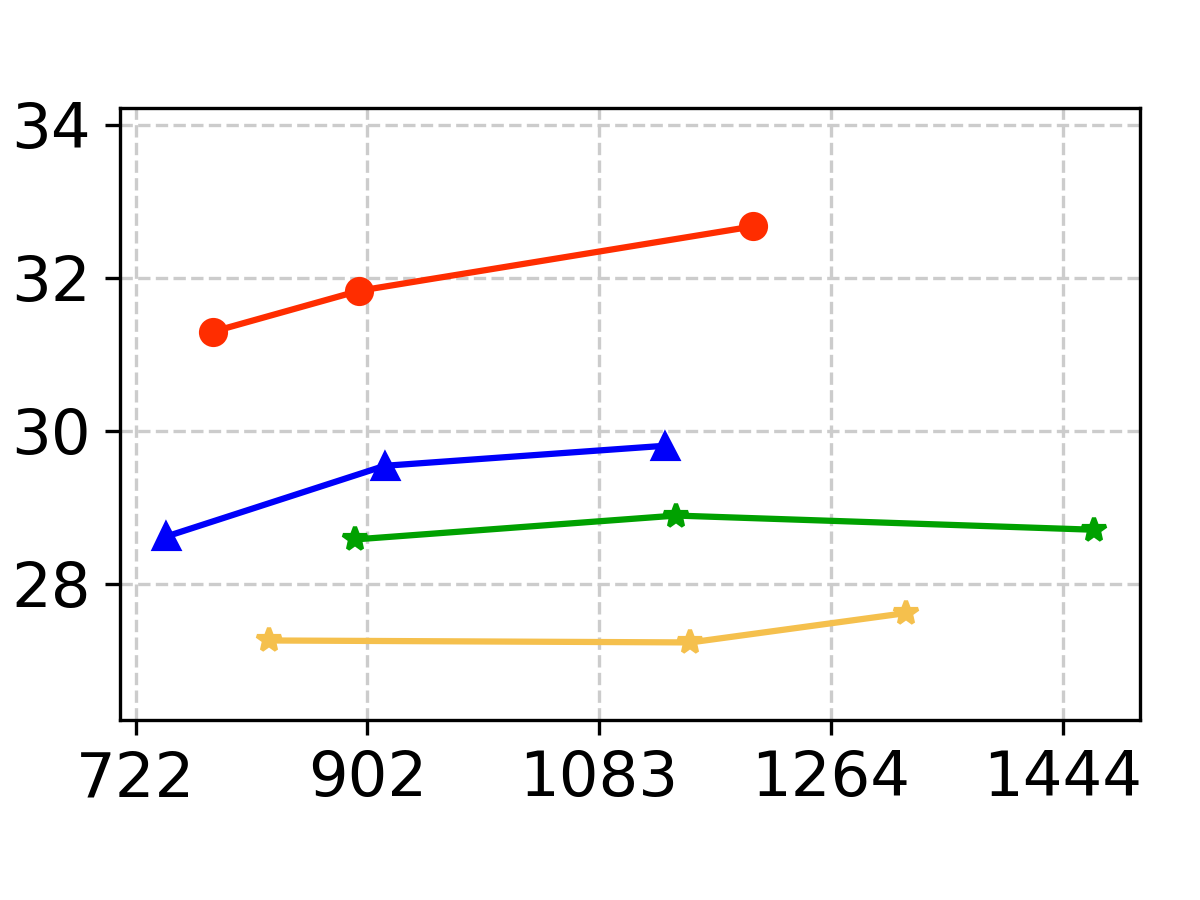}} & 
    \multirow{2}{*}[.6in]{\includegraphics[trim={0cm 0.5cm 0cm 0.5cm},clip,width=1.18\linewidth]{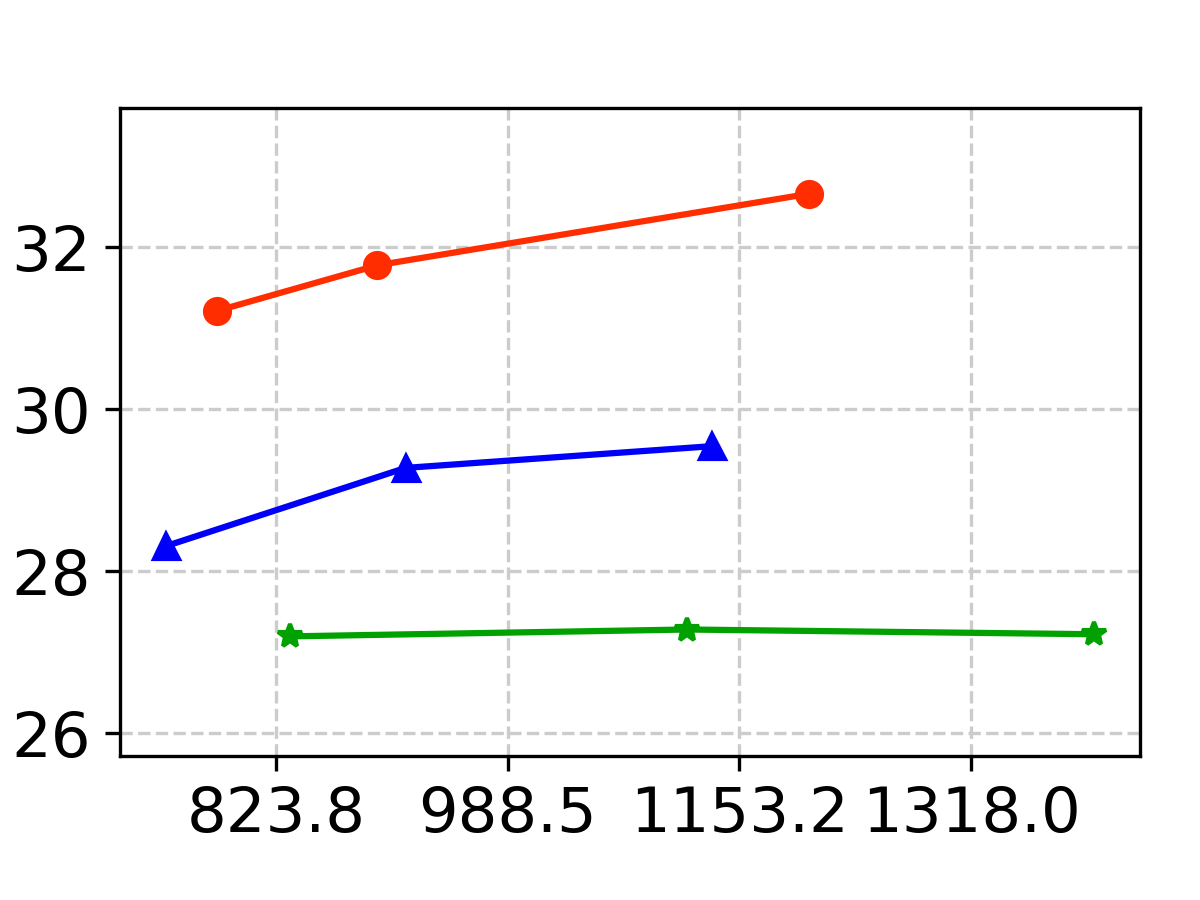}} &
    \multirow{2}{*}[.6in]{\includegraphics[trim={0cm 0.5cm 0cm 0.5cm},clip,width=1.18\linewidth]{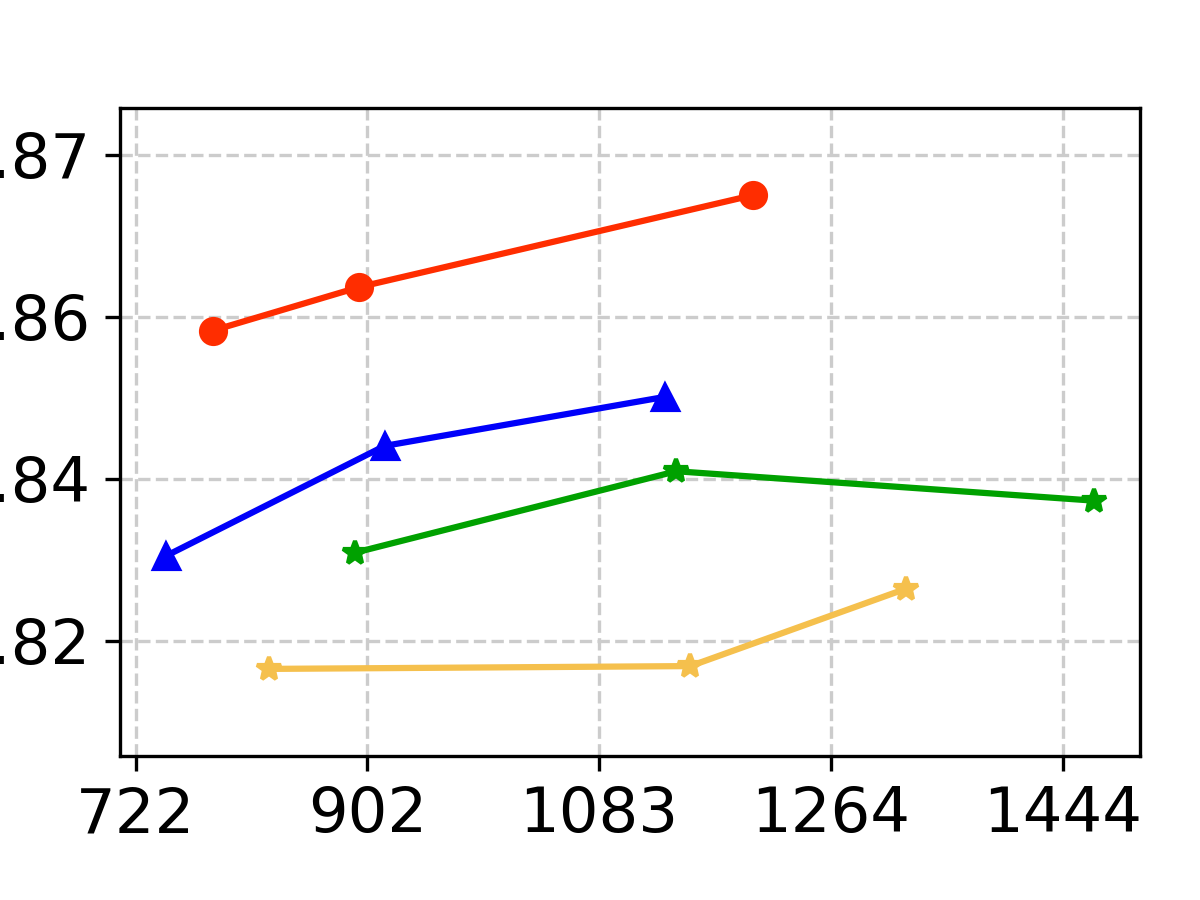}} &
    \multirow{2}{*}[.6in]{\includegraphics[trim={0cm 0.5cm 0cm 0.5cm},clip,width=1.18\linewidth]{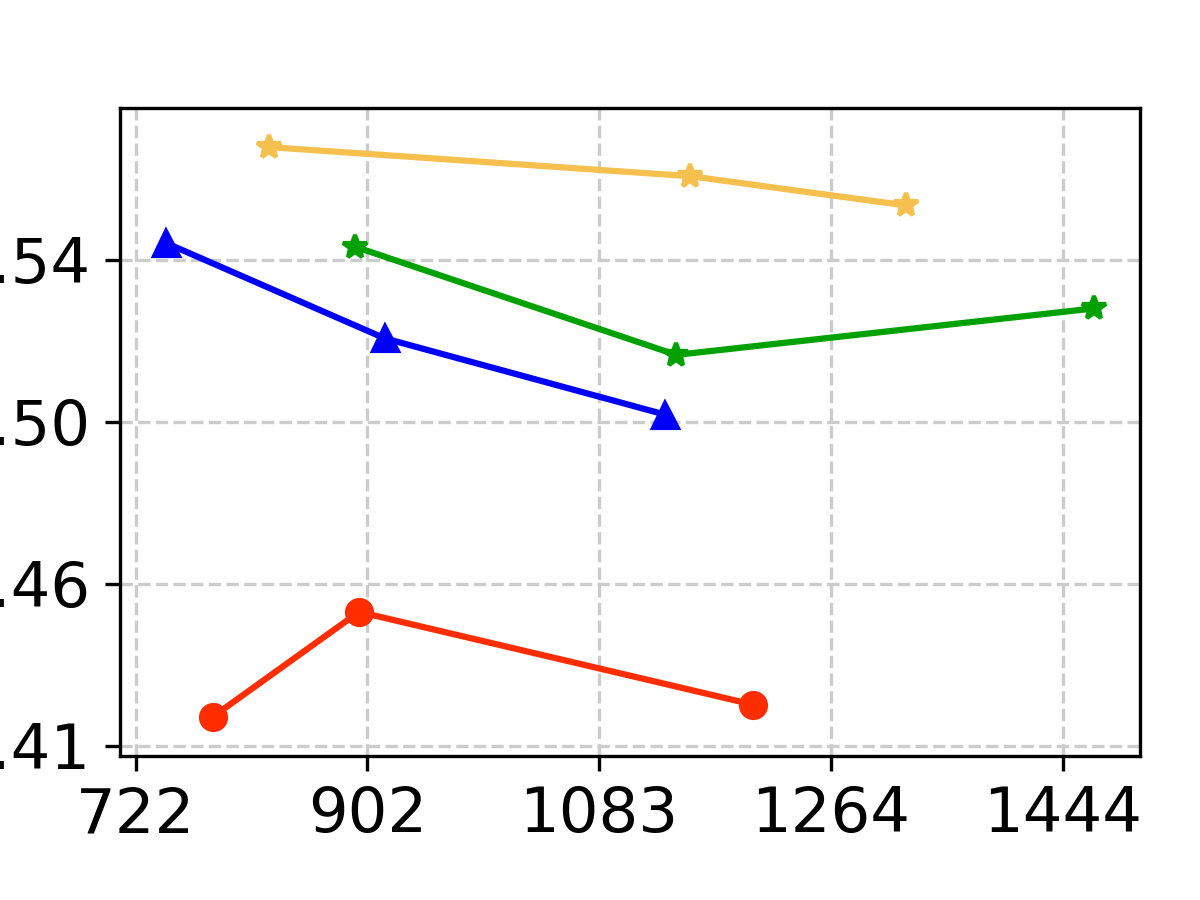}} \\
    \small{parrot} &  & & \\
    \\[-.8em]
    \includegraphics[trim={0cm 0cm 0cm 0cm},clip,width=1.1\linewidth]{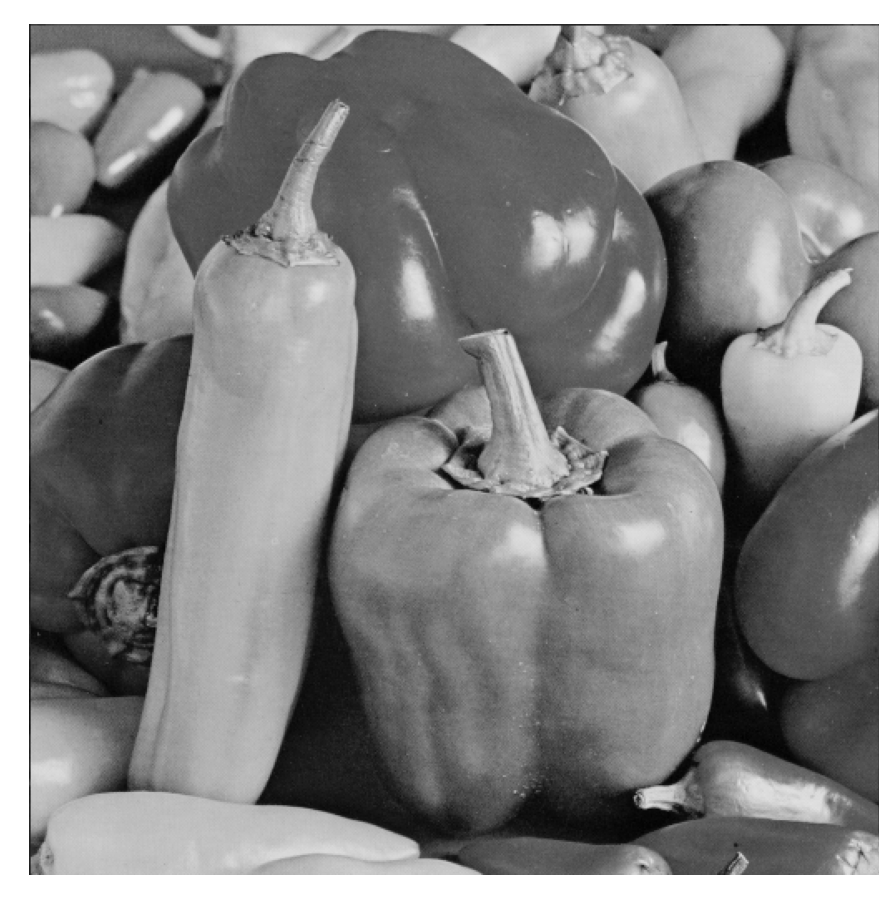} & 
     \includegraphics[trim={3cm 0.5cm 3cm 0cm},clip,width=1.1\linewidth]{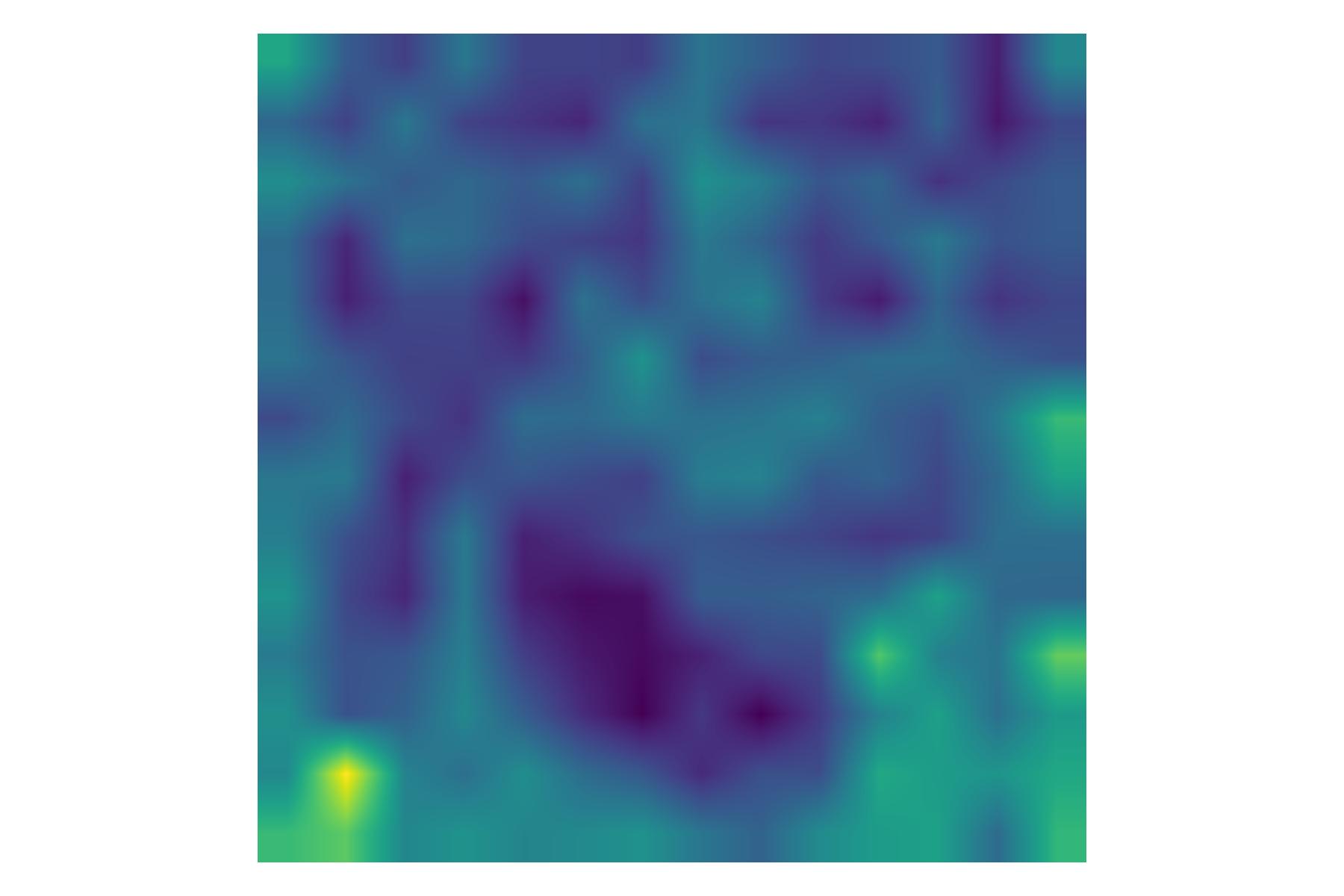} & 
    \multirow{2}{*}[.75in]{\includegraphics[trim={0cm 0cm 0cm 0.5cm},clip,width=1.18\linewidth]{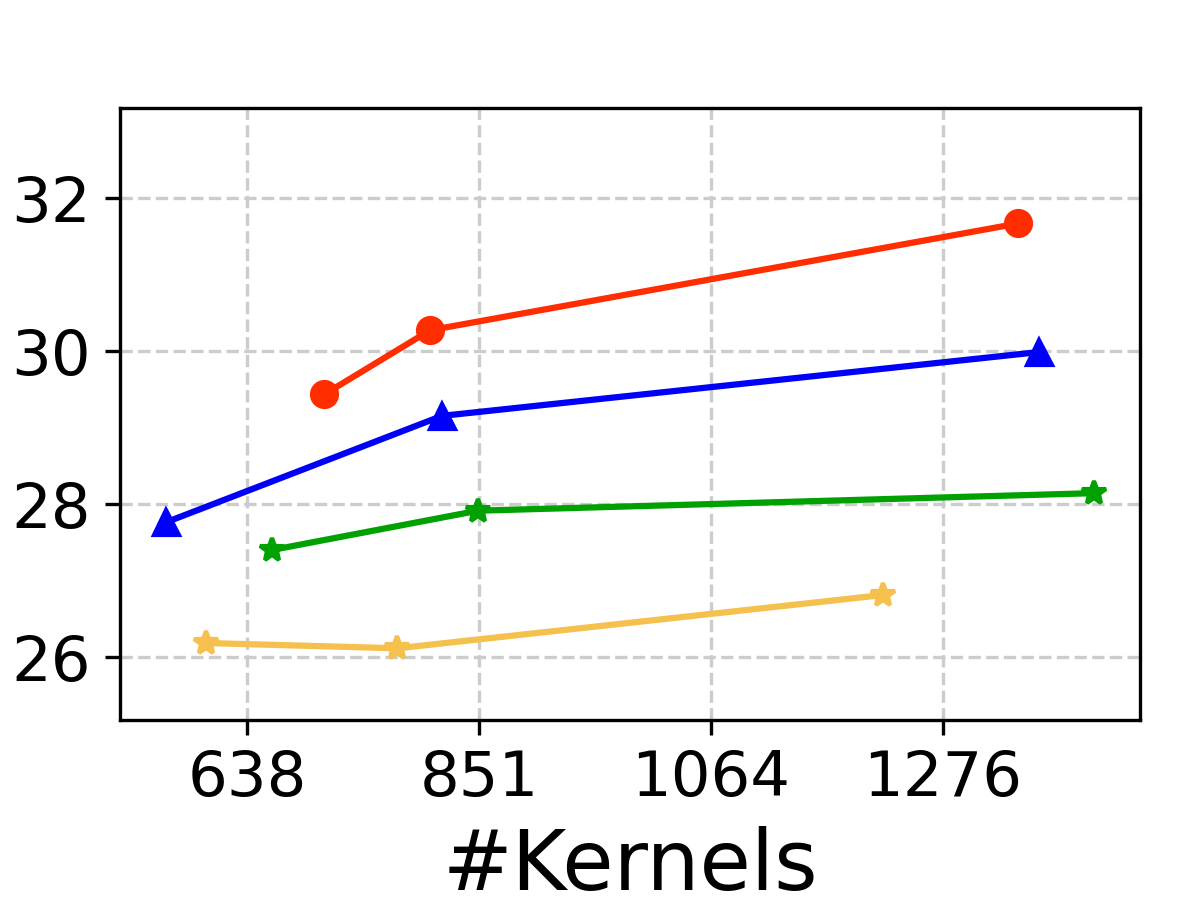}} & 
    \multirow{2}{*}[.75in]{\includegraphics[trim={0cm 0.5cm 0cm 0.5cm},clip,width=1.18\linewidth]{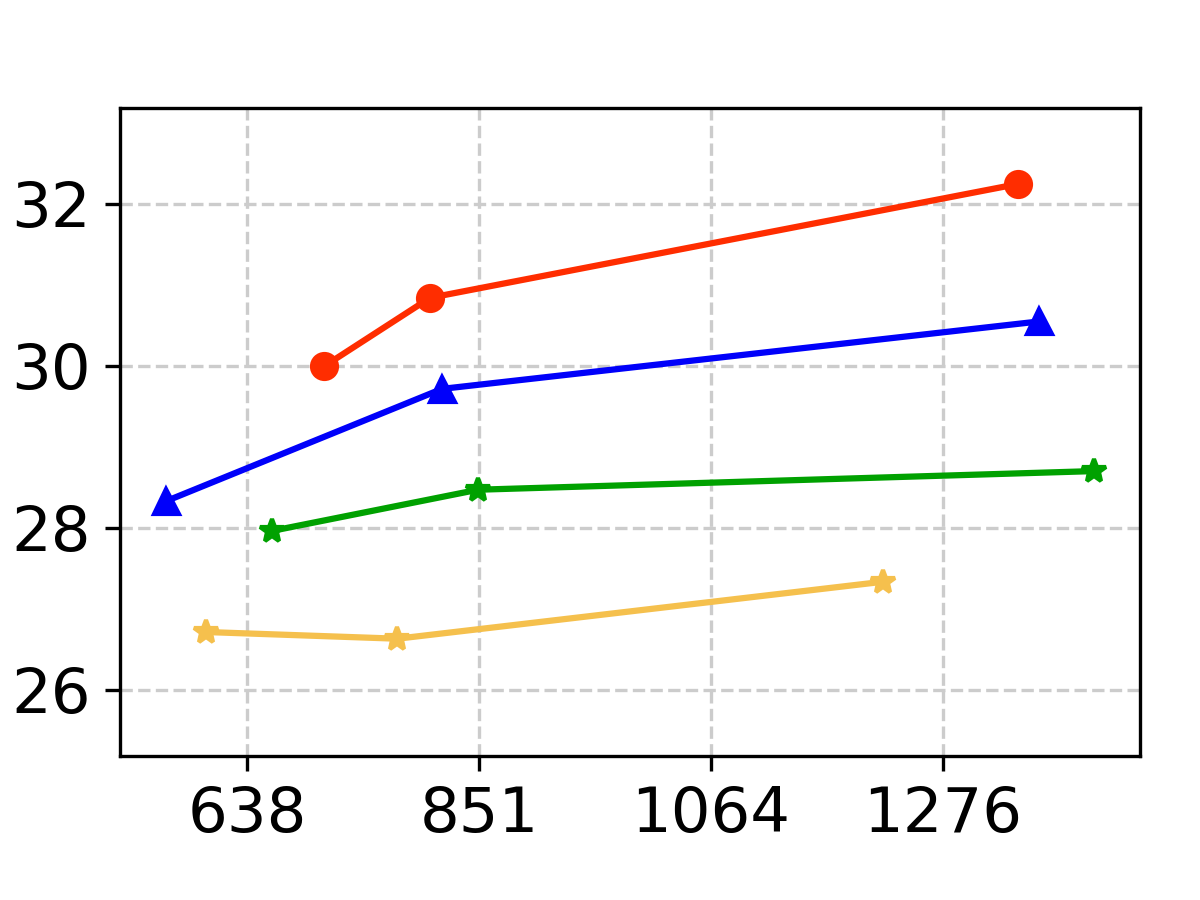}} &
    \multirow{2}{*}[.75in]{\includegraphics[trim={0cm 0.5cm 0cm 0.5cm},clip,width=1.18\linewidth]{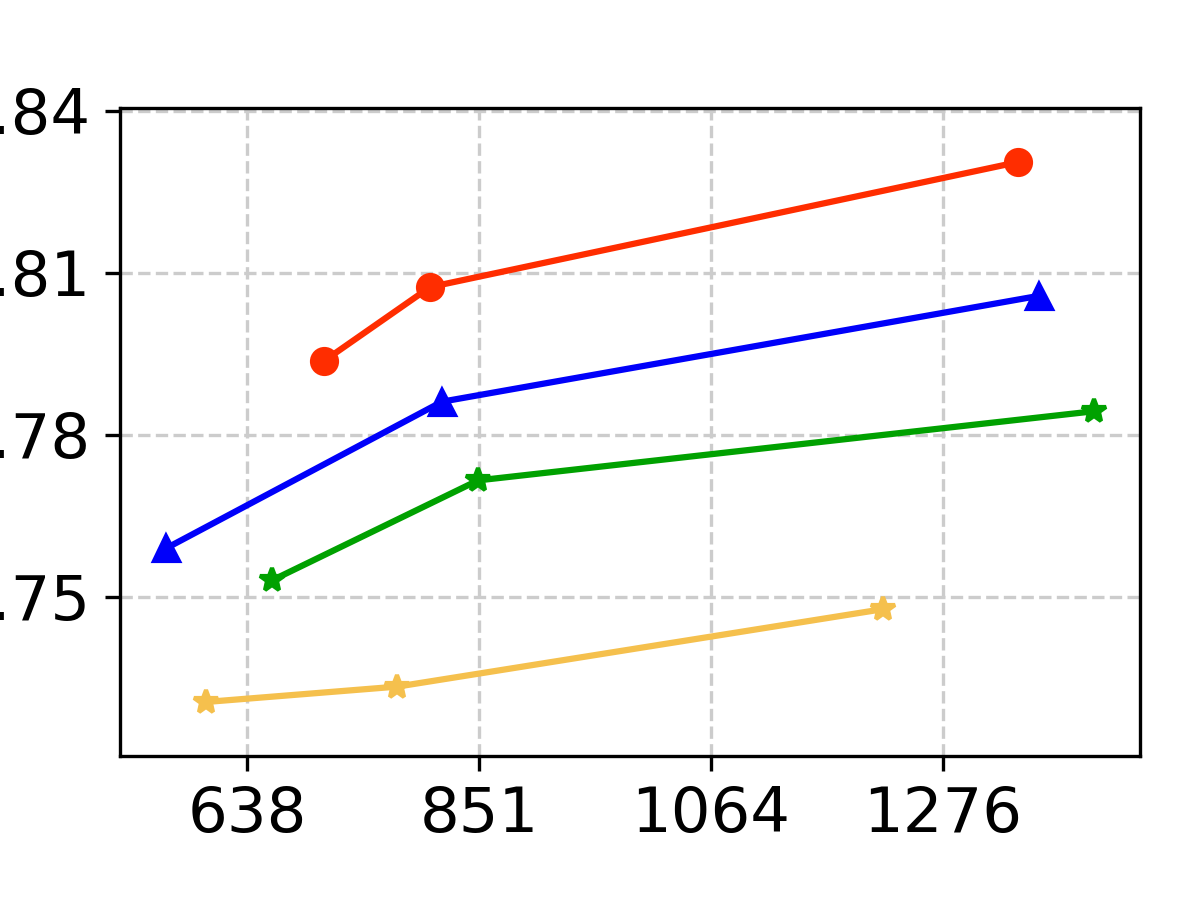}} &
    \multirow{2}{*}[.75in]{\includegraphics[trim={0cm 0.5cm 0cm 0.5cm},clip,width=1.18\linewidth]{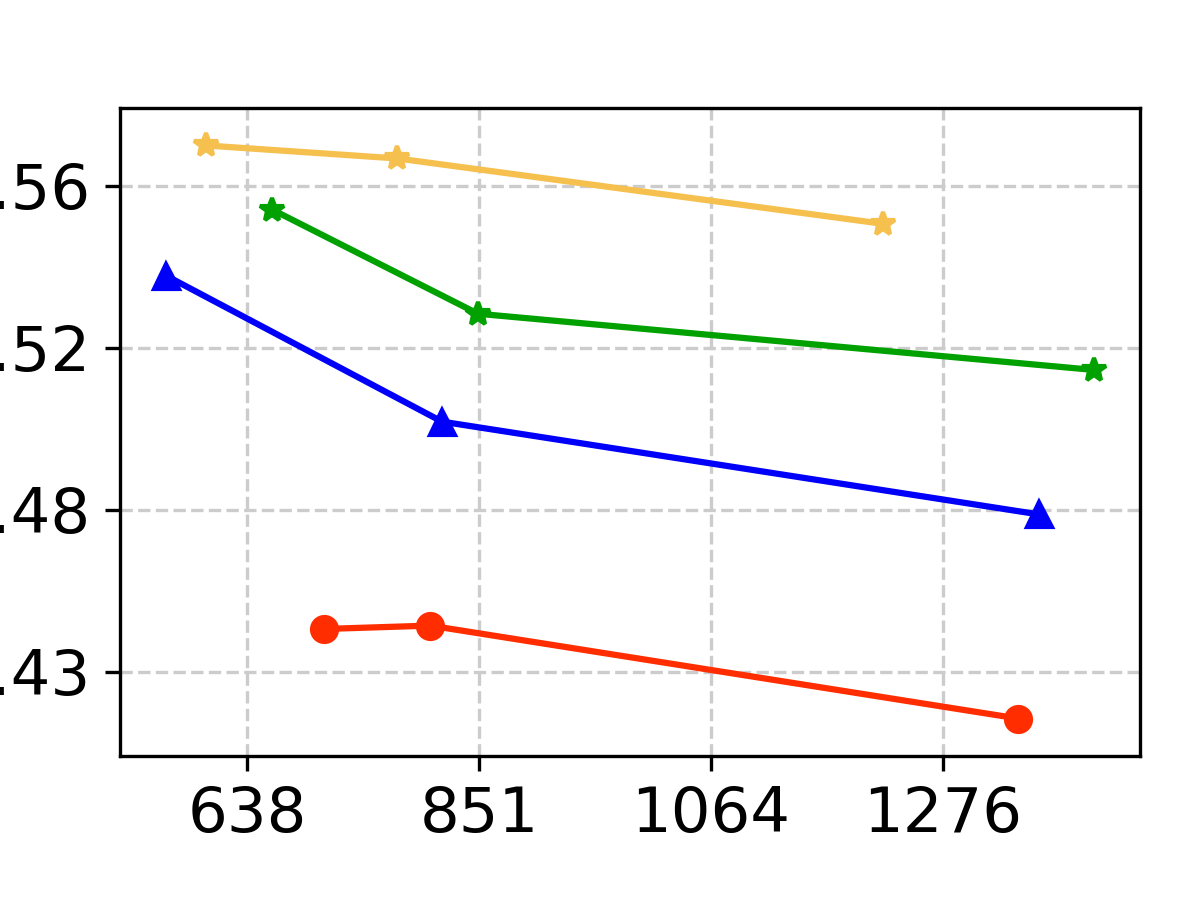}} \\
    [-0em]& & & \multicolumn{3}{c}{\includegraphics[trim={0cm 0cm 0cm 0cm},clip,width=0.47\linewidth]{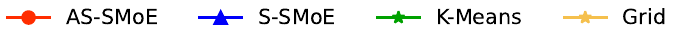}}\\
    \\
    [-3em]\small{peppers} & & & \\
    [-0em]& \small{attention map} & \small{ PSNR}& \small{ PSNR with attention} & \small{ SSIM} & \small{ LPIPS}
    
    \\
\end{tabularx}
\vspace{-1em}
\centering{\caption{Comparison of the reconstruction quality of Grid \cite{bochinski_regularized_2018}, K-Means \cite{verhack_steered_2020}, S-SMoE \cite{li_segmentation-based_2023}, and AS-SMoE}}
\vspace{-1em}
\label{fig5}
\end{figure*}

\begin{figure}[h]
\centering
\captionsetup{font=small}
\begin{tabularx}{0.5\textwidth} { 
   >{\centering\arraybackslash}X
   >{\centering\arraybackslash}X 
   >{\centering\arraybackslash}X 
   >{\centering\arraybackslash}X  }
   \small{Grid} & \small{K-Means} & \small{S-SMoE} & \small{AS-SMoE} \\
   \hline 
    \includegraphics[trim={.45cm 0.8cm 3.85cm 2.38cm},clip,width=2.cm]{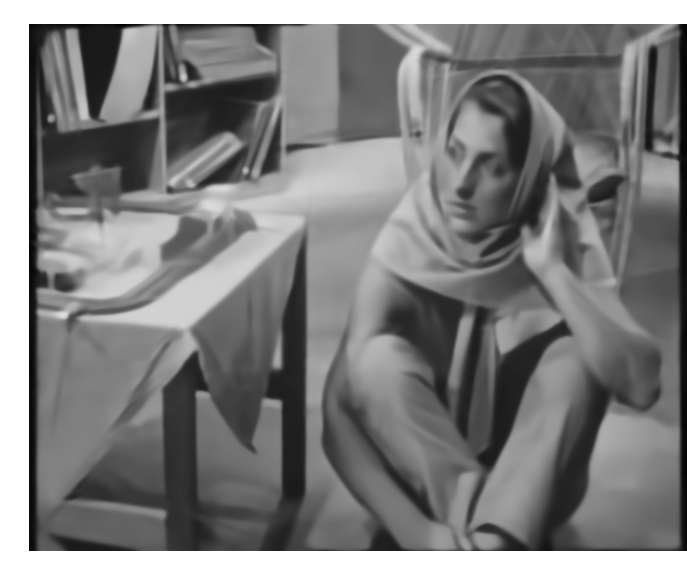} &
    \includegraphics[trim={.45cm 0.8cm 3.85cm 2.38cm},clip,width=2.cm]{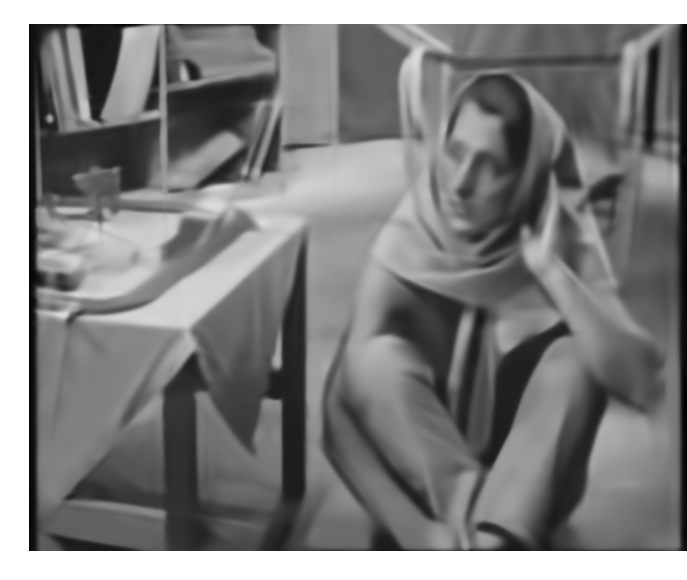} & 
    \includegraphics[trim={.5cm 1cm 5cm 3cm},clip,width=2.cm]{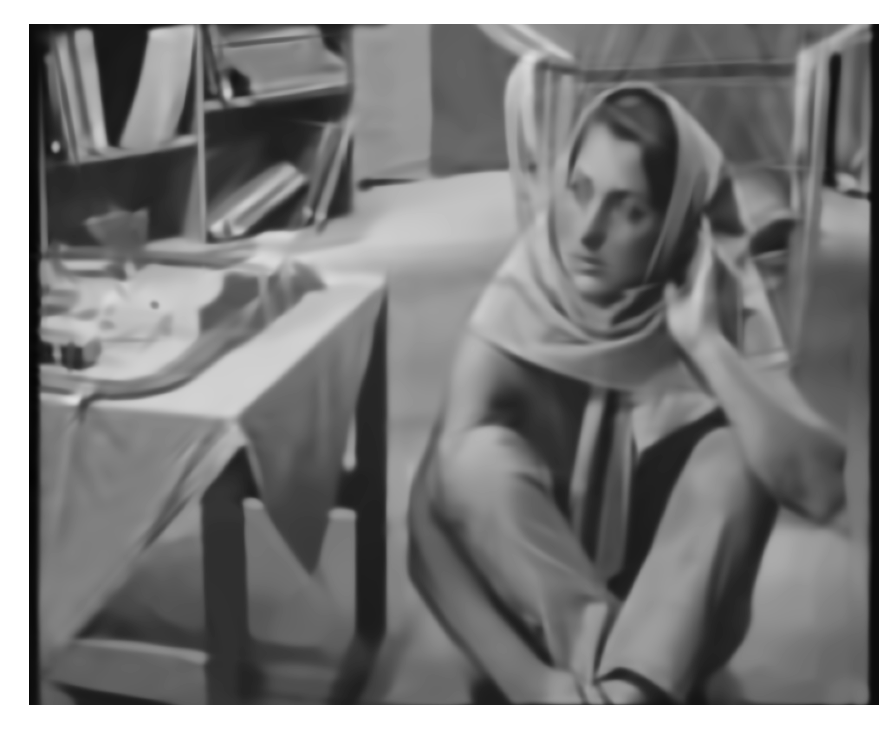} & 
    \includegraphics[trim={.5cm 1cm 5cm 3cm},clip,width=2.cm]{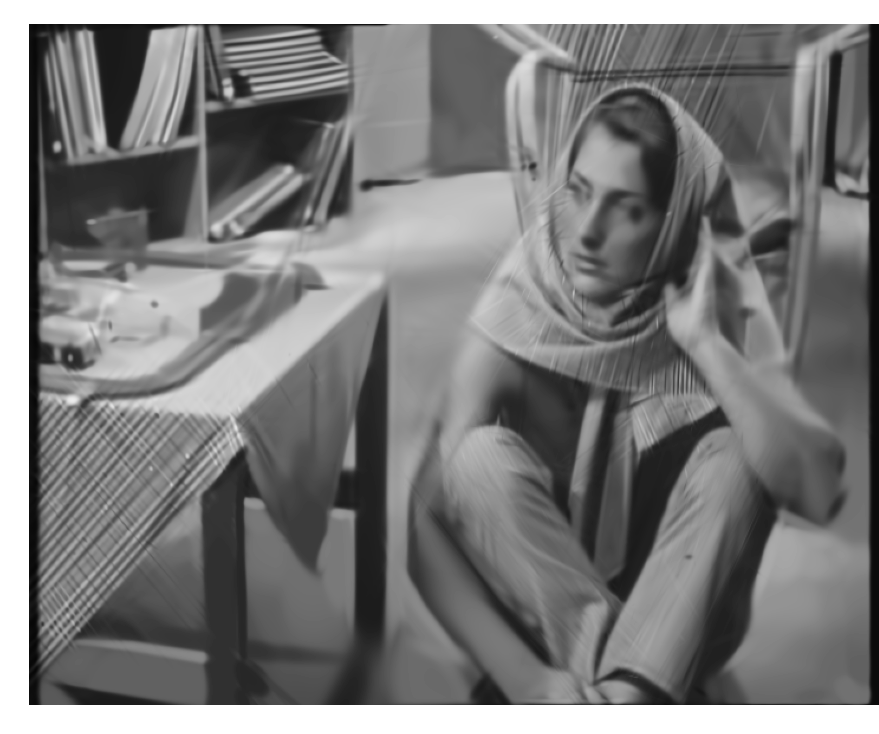} \\
    \small{L: 3279} & \small{L: 3186} & \small{L: 2824} & \small{L: 2922} \\
    \small{23.12 dB} & \small{22.71 dB} & \small{23.10 dB} & \small{24.66 dB} \\
    \includegraphics[trim={1.6cm 0.82cm 2.3cm .8cm},clip,width=2.cm]{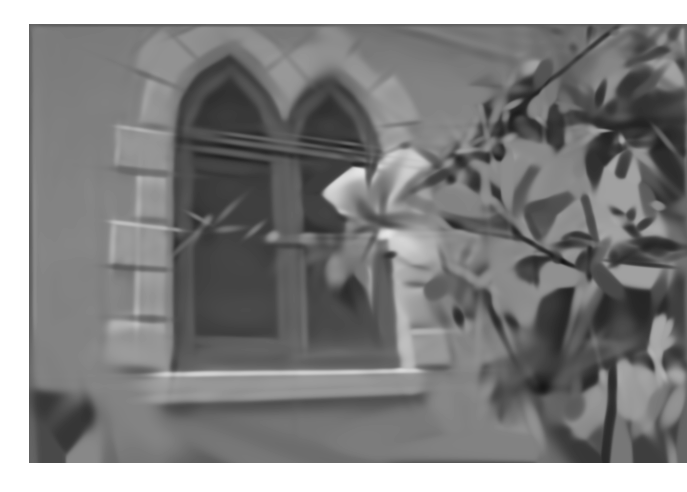} & 
    \includegraphics[trim={1.6cm 0.82cm 2.3cm .8cm},clip,width=2.cm]{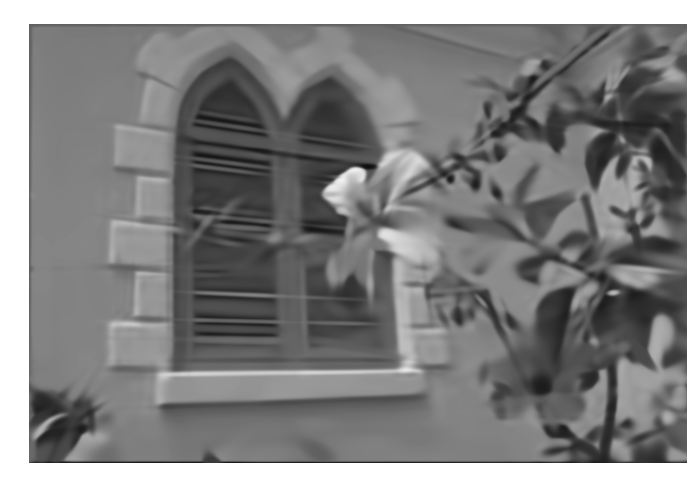} & 
    \includegraphics[trim={2cm 1cm 3cm 1cm},clip,width=2.cm]{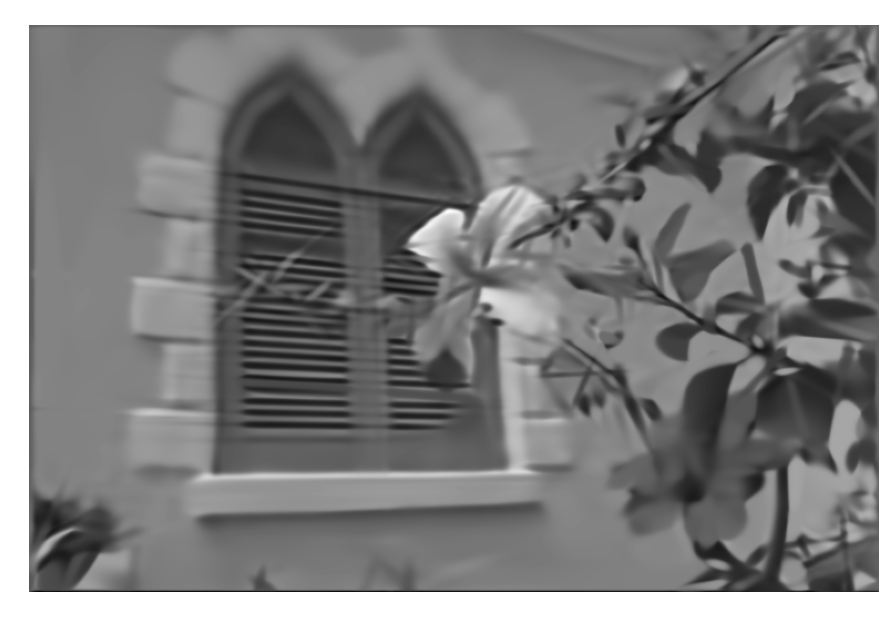} & 
    \includegraphics[trim={2cm 1cm 3cm 1cm},clip,width=2.cm]{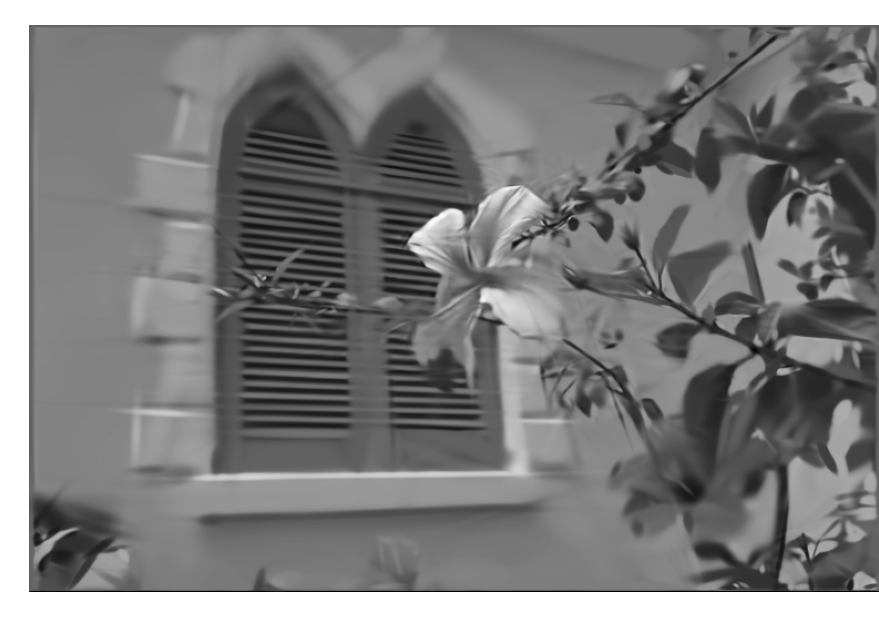} \\
    \small{L: 1101} & \small{L: 1334} & \small{L: 1060} & \small{L: 1178} \\
    \small{24.11 dB} & \small{25.19 dB} & \small{26.62 dB} & \small{28.54 dB} \\
    \includegraphics[trim={3.15cm 1.97cm 1.55cm 0.83cm},clip,width=2.cm]{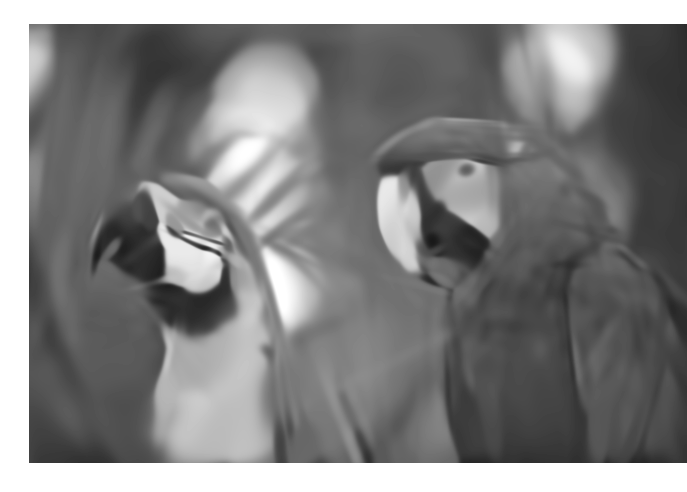} & 
    \includegraphics[trim={3.15cm 1.97cm 1.55cm 0.83cm},clip,width=2.cm]{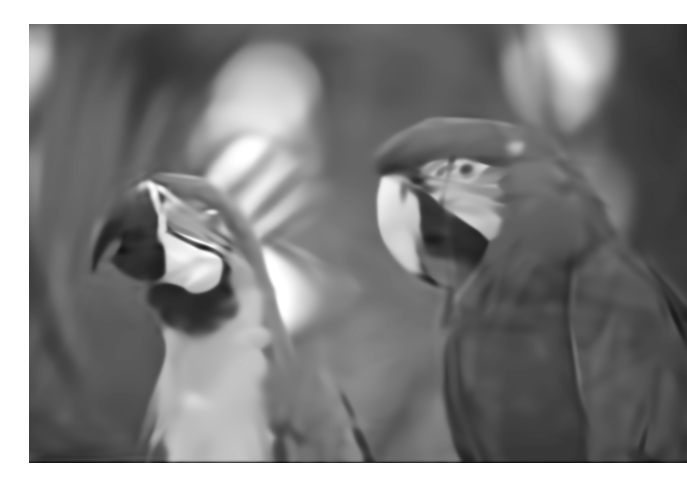} & 
    \includegraphics[trim={4cm 2.5cm 2cm 1cm},clip,width=2.cm]{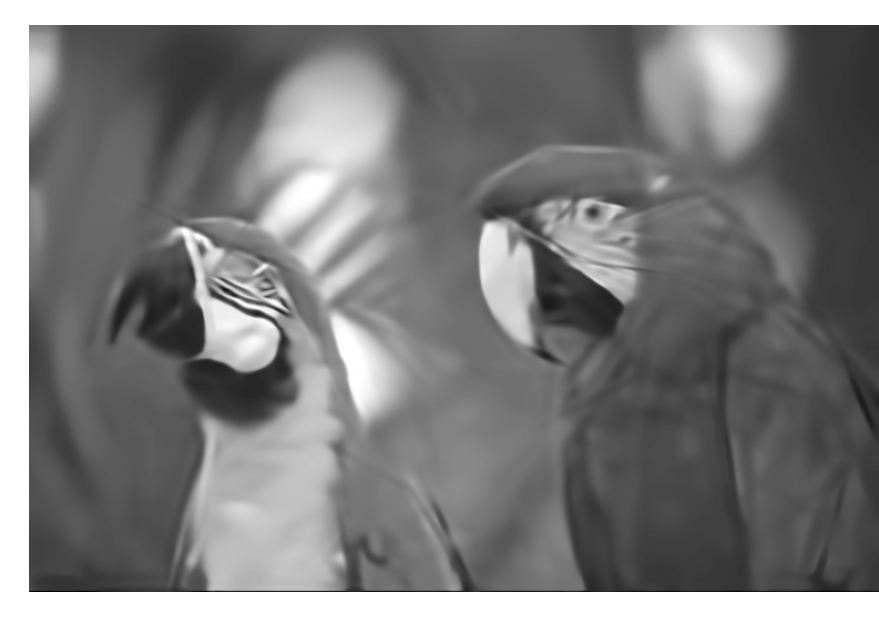} & 
    \includegraphics[trim={4cm 2.5cm 2cm 1cm},clip,width=2.cm]{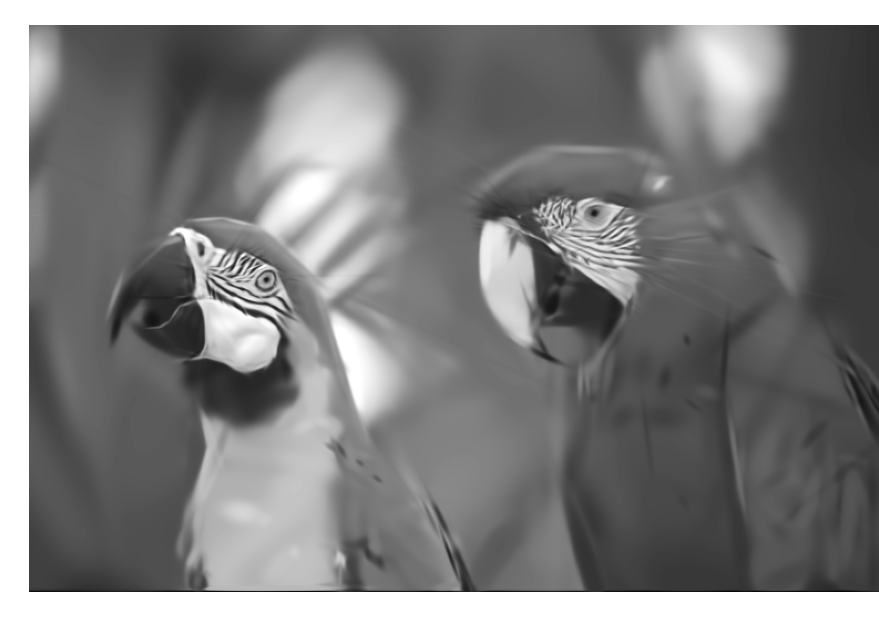} \\
    \small{L: 1322} & \small{L: 1468} & \small{L: 916} & \small{L: 896} \\
    \small{27.23 dB} & \small{28.31 dB} & \small{29.13 dB} & \small{31.39 dB} \\
\end{tabularx}
\vspace{-.5em}
\caption{Visualization of the high-frequency AS-SMoE reconstruction in comparison to Grid, K-Means, and S-SMoE ($d_{Th}$=40). PSNR performance is displayed below each kernel count L.}
\vspace{-2em}
\label{fig6}
\end{figure}

\section{Results}

Table \ref{table3} and Fig.5 provide insight into the performance gains of segmentation-based initialization methods S-SMoE and AS-SMoE against Grid and K-Means on the test set. MDBSCAN segmentation pixel difference thresholds $d_{Th}$ values 40 and 50 were investigated to evaluate the impact of the segmentation details. While Table \ref{table3} provides emphasis on the reconstruction quality, Fig. 5 allows to evaluate the gains in a Rate-Distortion or Sparsity-vs-Quality context. The results shown refer to the reconstruction quality after final Global Optimization (the last step in Fig. \ref{fig2}). The Global Optimization algorithm used again a GD optimization with kernel sparsification, initialized with the three initialization strategies. In Eq. (\ref{eq:6}) the loss is now evaluated on all pixels of the respective test image. $S$ now is the whole image, and $L$ is in the sum of all kernels from the merging, which is then adaptively adjusted. The global SMoE model regression reconstruction $y_p(x)$ starts with the initial number of $L_{Init}$ of kernels provided by the global initialization steps. During Global Optimization, the number of kernels decreases adaptively as the sparsity parameter $\pi$ drops below zero, until the decrease in losses between iterations falls below a threshold of $10^{-6}$. With our results, we provide the effective number of kernels $L_{Opt}$, which occurs after global optimization with the sparsity parameter $\pi$ greater than zero. $L_{Opt}$ is an important measure of SMoE model complexity or sparsity of the model. The computational complexity and memory required to reconstruct the images are linearly dependent on $L_{Opt}$. For many applications, such as SMoE image compression, the number of kernels is also an estimate of the entropy. The fewer kernels for a given reconstruction quality, the fewer bits to code. 


We can see in Table \ref{table3} that both S-SMoE and AS-SMoE consistently outperform the established Grid and K-Means initialization by significant numbers. This is consistently case for all quality measures evaluated, as well as for more or less textured segmentation using different $d_{Th}$. In particular, the AS-SMoE approach, which extends the non-adaptive segmentation S-SMoE strategy with adaptive initialization, provides again significantly improved quality of reconstruction. Compared to Grid and K-Means, the proposed AS-SMoE method demonstrates a drastic PSNR gain of up to more than 4 dB for the test images. Over S-SMoE initialization again a very significant improvement between 1 and 2.2 dB is realized, depending on the test image. The quality gains appear also well reflected in the SSIM and LPIPS measures, even though we optimized the objective function $L$ towards MSE rather than towards SSIM and LPIPS measures, which would have also been possible using the Global Optimization step. 

From Fig. 5 we observe that more kernels $L_{Opt}$ in general result in better reconstruction quality for a given initialization method, independent of the quality measure employed. The exception is test image Barbara, where obviously the delicate fine-grained texture details are not, or not sufficiently, improved with more kernels for S-SMoE and AS-SMoE. Notice, that the S-SMoE and AS-SMoE improve the reconstruction quality while simultaneously arriving in sparser representations with fewer kernels compared to Grid and K-Means. This result makes the AS-SMoE initialization highly attractive beyond objective reconstruction quality considerations. The drastic objective quality improvements of S-SMoE and AS-SMoE also translate into impressive subjective quality gains. Fig. \ref{fig6}  shows crops of reconstructed images from the test set. The visual quality of reconstructed images drastically improves from Grid and K-Means to S-SMoE towards AS-SMoE. The AS-SMoE initialization results in high-frequency texture content being reconstructed efficiently, while Grid, K-Means, and S-SMoE provide far less satisfactory details. Notice that this improvement is achieved even with significantly fewer kernels $L_{Opt}$ after optimization - the SMoE models improve significantly their explain-ability and sparsity. 

We illustrate the sparsity enforcing capability of AS-SMoE under various controlled conditions. The Global Optimization process was terminated upon reaching the target PSNR reconstruction quality. This target quality varied depending on the complexity of the individual testing images. This allows a comparison of the model complexity with reconstructed test images at the same PSNR. Table \ref{table4} compares the number of kernels $L_{Opt}$ of the SMoE models finally used for the reconstruction of the images. The results show that the proposed AS-SMoE initialization strategy resulted in models with on average 49\%, 54\%, and 41\% reduced number of kernels compared to Grid, K-Means, and S-SMoE initialization strategies, respectively. This appears to be a remarkable sparsity gain. 

\begin{figure}[t]
\centering
\captionsetup{font=small}
\begin{tabularx}{0.5\textwidth} { 
   >{\centering\arraybackslash}X 
   >{\centering\arraybackslash}X  }
    \includegraphics[trim={1.9cm 0.2cm 1cm 0cm},clip,width=4.3cm]{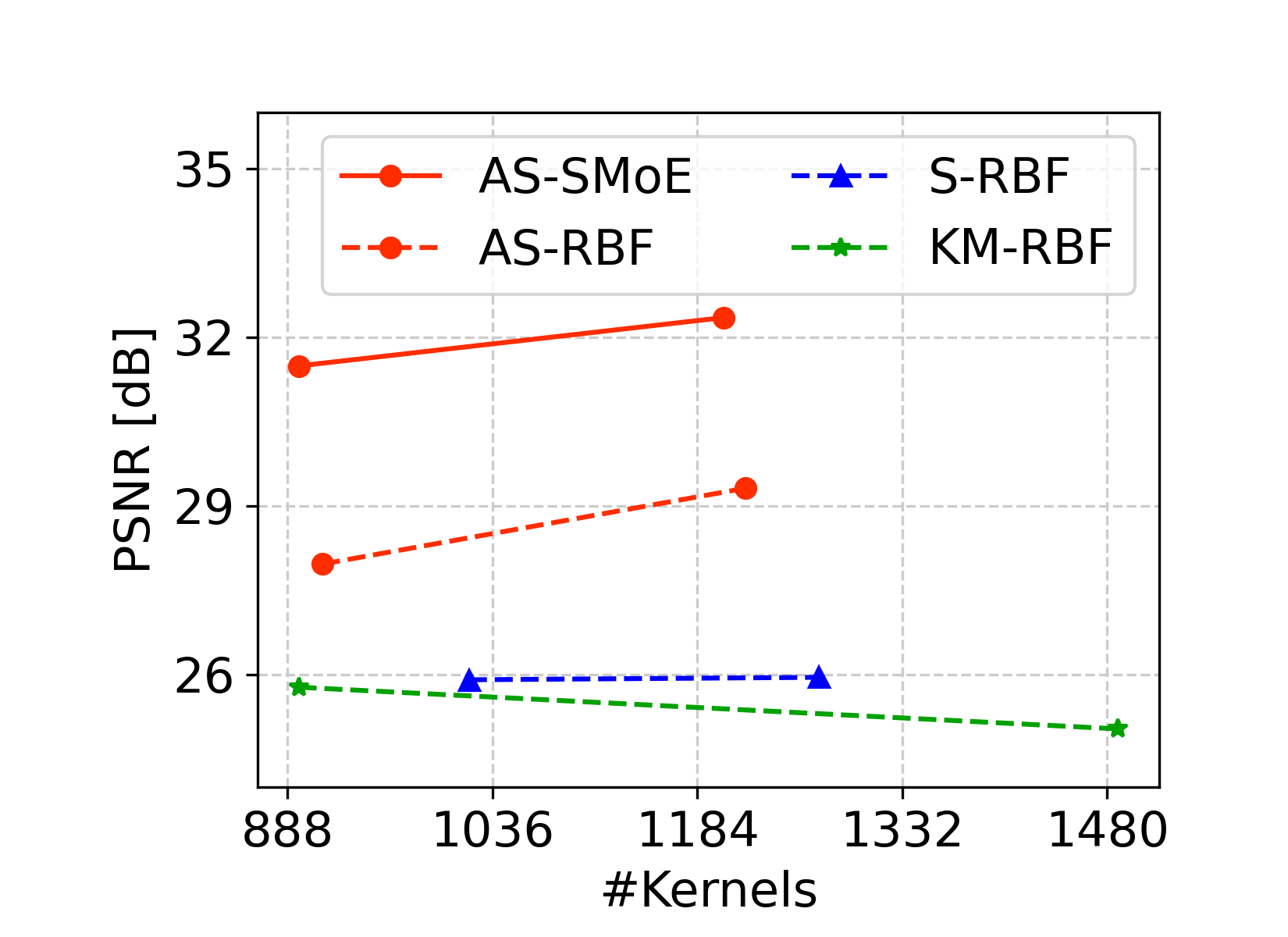} & 
    \includegraphics[trim={3cm 1cm 0cm .5cm},clip,width=4cm]{figures/recon/40_proposed_25_parrot.png} \\
    \small{PSNR comparison} & \small{AS-SMoE, PSNR: 31.39 dB} \\
    \includegraphics[trim={3cm 1cm 0cm .5cm},clip,width=4cm]{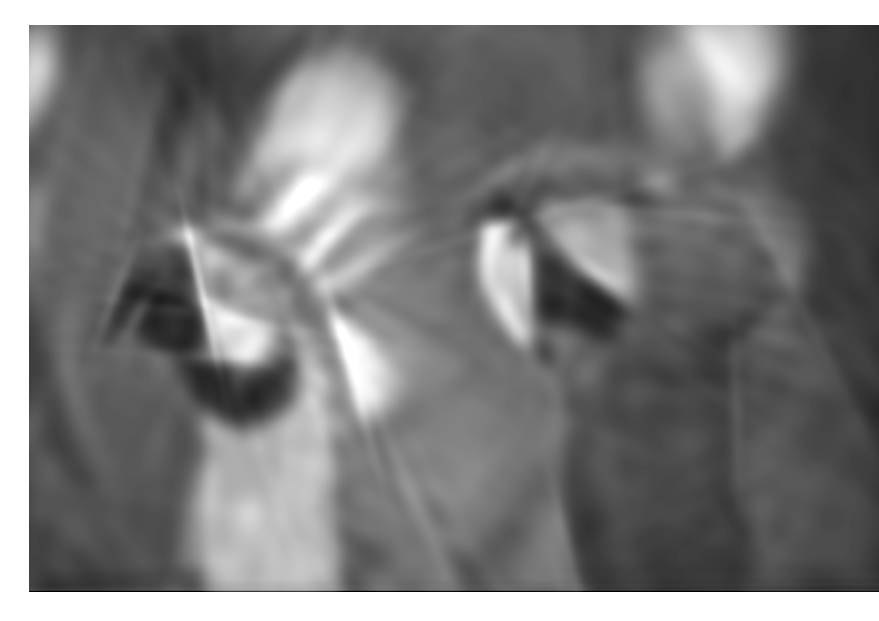} & 
    \includegraphics[trim={3cm 1cm 0cm .5cm},clip,width=4cm]{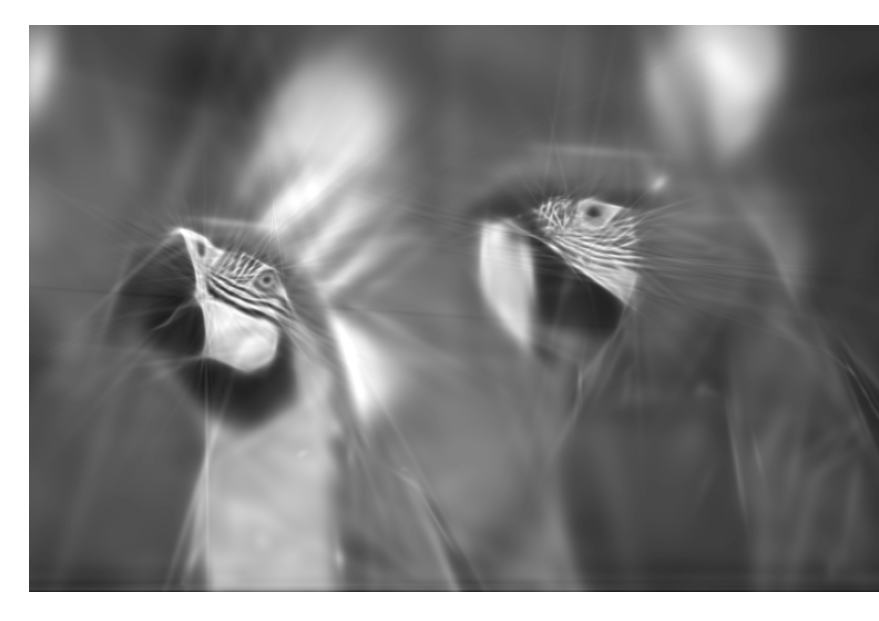} \\
    \small{KM-RBF, PSNR: 25.87 dB} & \small{AS-RBF, PSNR: 27.98 dB} \\
\end{tabularx}
\caption{Comparison of two initialization methods (K-Means and AS) with RBF and SMoE regression.}
\vspace{-1em}
\label{fig7}
\end{figure}

\begin{table}[t]
\captionsetup{font=small,justification=centering, labelsep=newline,textfont=sc}
\vspace{1em}
\caption{The number of kernels required to achieve reconstruction quality target in PSNR ($d_{Th}=40$)}
\begin{tabular}{>{\centering}m{.025\textwidth}>{\centering}m{.03\textwidth}|>{\centering}m{.028\textwidth}|>{\centering}m{.035\textwidth}|>{\centering}m{.035\textwidth}|>{\centering}m{.015\textwidth}>{\centering}m{.035\textwidth}>{\centering}m{.03\textwidth}>{\centering\arraybackslash}m{.035\textwidth}}
\hline
         &              & {Grid} & {K-Means} & {S-SMoE} & \multicolumn{4}{c}{AS-SMoE} \\
         \cline{3-9}
Image &  PSNR & \multirow{2}{*}{$L_{Opt}$} & \multirow{2}{*}{$L_{Opt}$} & \multirow{2}{*}{$L_{Opt}$} & \multirow{2}{*}{$L_{Opt}$} & \multicolumn{3}{c}{Kernel reduction vs.} \\
         &              &           &     &      &           &  Grid & K-Means & S-SMoE\\[0ex]
\hline
{barbara} &25 &4308 &4265 & 4247 & 1653 &62\% &61\% & 61\%\\
{flower}  &26 &1619 &1956 & 1060 & 807 &50\% &59\%  & 24\% \\
{parrot}  &30 &1322 &1468 & 1134 & 782 &41\% &47\%  & 31\% \\
{peppers} &29 &1222 &1415 & 1365 & 709 &42\% &50\%  & 49\%  \\
\hline
\multicolumn{6}{l}{Average}         &48.75\% &54.25\%  & 41.25\% \\
\hline
\end{tabular}
\vspace{-2em}
\label{table4}
\end{table}

While the prime focus of the paper is to show the gains of improved segment-based initialization for SMoE gating network models, it is beneficial to investigate the novel initialization strategy in the context of Radial Basis Function Networks. Results in Fig.  \ref{fig7} show, that RBF models also significantly benefit from the proposed segmentation-based initialization (AS-RBF). Compared to K-Means (KM-RBF) an impressive PSNR gain of 2.1 dB also results in drastically improved visual improvement. Initialized with the non-adaptive segmentation-based strategy (S-RBF) does not provide benefits for RBF regression. Most importantly, however, the SMoE gating networks (AS-SMoE) significantly outperform the steered RBF regression. A drastic improvement of appr. 3.4 dB again over the AS-RBF model results in significantly improved visual quality. The edge-preserving capability of SMoE gating models particulary improves the reconstruction of edges and lines in the images, as expected. 


The way how parameters of different initialization methods impact on quality and sparsity of the optimized SMoE models differs for the individual initialization methods - and is not straight-forward to analyze. Grid, K-Means, and S-SMoE initialization methods generate round (non-steered) Gaussian kernels. The locations of these kernels with identical bandwidth are used for Global Optimization. Although the steered parameters are trained during global optimization, they remain undisclosed in the initialization stage. In contrast, AS-SMoE produces locations of steered kernels, each with individual steering and bandwidth properties derived from the kernel covariance matrices. The locations and the covariance matrix parameters are used for Global Optimization. The better the initialization reconstruction of the SMoE model in the first iteration with highly suitable initial kernel parameters, the faster the convergence, the better the final reconstruction quality, and the sparser the model. Table \ref{table5} shows the initialization quality of SMoE models. Results were generated from the initial number of kernels $L_{Init}$ and their parameter setting without Global Optimization ($d_{Th}=40$). This complies to the first initial step in GD optimization. The results confirm, that S-SMoE and AS-SMoE initialization consistently provide significantly improved initial reconstruction quality for subsequent optimization, with best improvements obtained with the AS-SMoE approach. Notice, that the S-SMoE and AS-SMoE initialization methods generated significantly fewer kernels for reconstruction compared to both Grid and K-Means - approx. 17\% fewer kernels for S-SMoE and 15\% reduction of kernels for AS-SMoE for test image Barbara.

\begin{figure*}[t]
\centering
\captionsetup{font=small}
\subfloat[\textit{Grid, 14.98dB}]{\includegraphics[width=.20\linewidth]{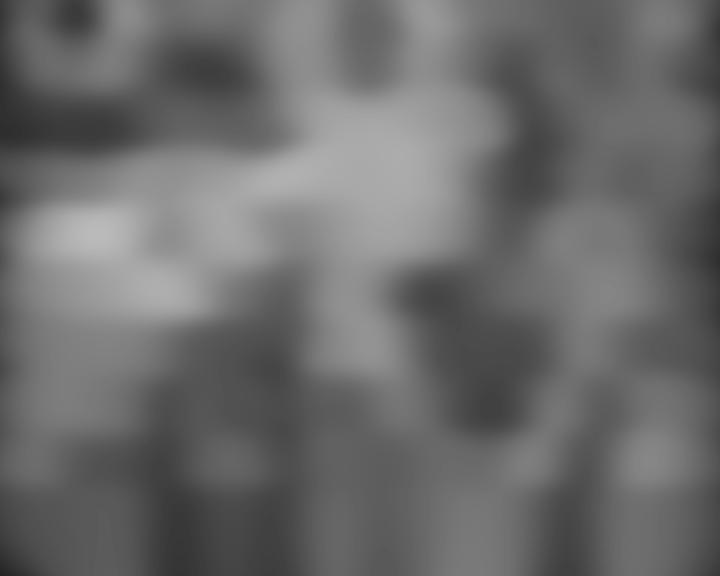}}
\subfloat[\textit{K-Means, 14.46dB}]{\includegraphics[width=.20\linewidth]{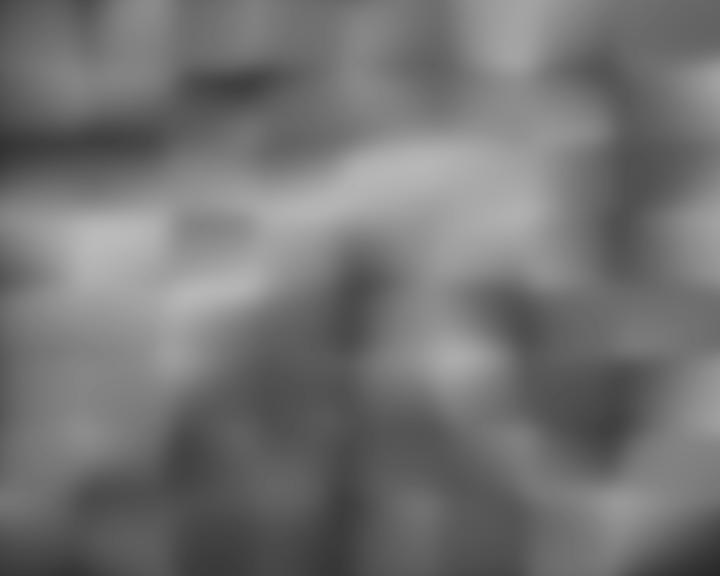}}
\hfil
\subfloat[\textit{Segmentation}]{\includegraphics[width=.20\linewidth]{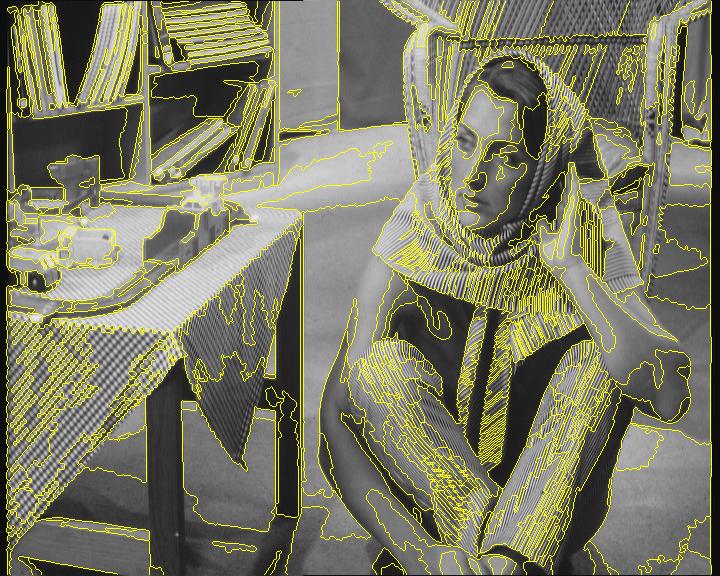}}
\hfil
\subfloat[\textit{S-SMoE, 15.19dB}]{\includegraphics[width=.20\linewidth]{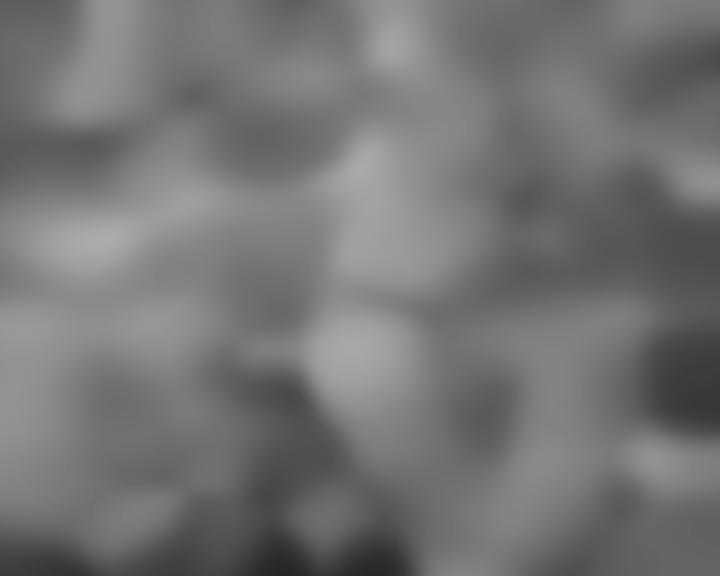}}
\hfil
\subfloat[\textit{AS-SMoE, 15.87dB}]{\includegraphics[width=.20\linewidth]{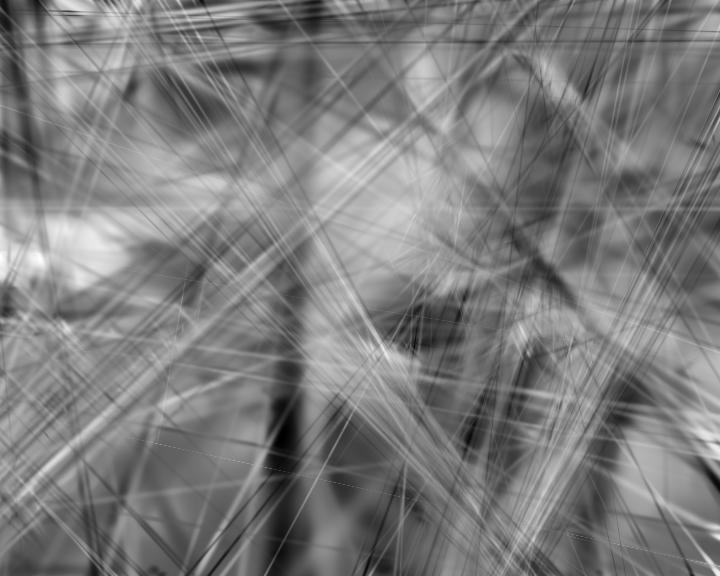}}
\caption{Comparison of the initialization results with Grid, K-Means, S-SMoE, and AS-SMoE for $d_{Th}=40$. The MDBSCAN segmentation result for S-SMoE and AS-SMoE is shown in (b).}
\vspace{-1em}
\label{fig8}
\end{figure*}

\begin{table*}[t]
\captionsetup{font=small,justification=centering, labelsep=newline,textfont=sc}
\vspace{1em}
\caption{Initialization Quality}
\begin{tabular}{>{\centering}m{.04\textwidth}|>{\centering}m{.04\textwidth}>{\centering}m{.08\textwidth}>{\centering}m{.045\textwidth}|>{\centering}m{.04\textwidth}>{\centering}m{.08\textwidth}>{\centering}m{.045\textwidth}|>{\centering}m{.04\textwidth}>{\centering}m{.08\textwidth}>{\centering}m{.045\textwidth}|>{\centering}m{.04\textwidth}>{\centering}m{.08\textwidth}>{\centering\arraybackslash}m{.025\textwidth}}
\hline
  & \multicolumn{3}{c|}{Grid} & \multicolumn{3}{c|}{K-Means} & \multicolumn{3}{c|}{S-SMoE} & \multicolumn{3}{c}{AS-SMoE} \\
\hline
Image &  $L_{Init}$&PSNR [dB] & SSIM &  $L_{Init}$&PSNR [dB] & SSIM & $L_{Init}$&PSNR [dB] & SSIM & $L_{Init}$&PSNR [dB] & SSIM \\
\hline
\multirow{1}{*}{barbara}&3500&14.98&0.34&3500 & 14.46	&0.35	&2928 &15.19  &0.35  &2991 &15.87  &0.35\\
\multirow{1}{*}{flower} &1500&16.27&0.51&1500	&15.56	&0.50   &1165 &17.05  &0.50  &1206 &18.49  &0.50\\
\multirow{1}{*}{parrot} &1500&14.40&0.65&1500	&13.68	&0.64   &1019 &16.27  &0.65  &913  &16.37  &0.62\\
\multirow{1}{*}{peppers}&1500&11.73&0.44&1500	&15.12	&0.48   &864 &14.49  &0.47  &811 &16.92  &0.53\\
\hline
\end{tabular}

\vspace{-1em}
\label{table5}
\end{table*}

Fig. \ref{fig8} allows visual inspection of the result in Table \ref{table4} for test image Barbara. S-SMoE achieves an initialization gain over Grid and K-Means of approximately 0.2 and 0.7 dB, respectively. This results in approximated 0.3 and 0.7 dB gain after Global Optimization (c.f. Table \ref{table2}). AS-SMoE results in a significantly improved 1.3 dB initialization gain compared to Grid and K-Means, which converts into 1.5 and 1.9 dB improvements on the optimized reconstruction. The contours of ``Barbara" are now visible in the initialization image of AS-SMoE. High frequency details, such as edges are present. The apparently disturbing sharp and elongated steering kernels visible in the image are not artifacts. They provide important initial steering information, which allows the SMoE Global Optimization to converge to a more accurate and more sparse model with reduced numbers of iterations. This is apparent by inspecting the legs of ``Barbara". The parallel small-width steered kernels already represent the patterns of the trousers - on both legs steering in different directions.

\begin{figure*}[t]
\centering
\captionsetup{font=small}
\subfloat[\textit{barbara}]{\includegraphics[trim={0.2cm 1cm 2.1cm 2cm},clip,width=.33\linewidth]{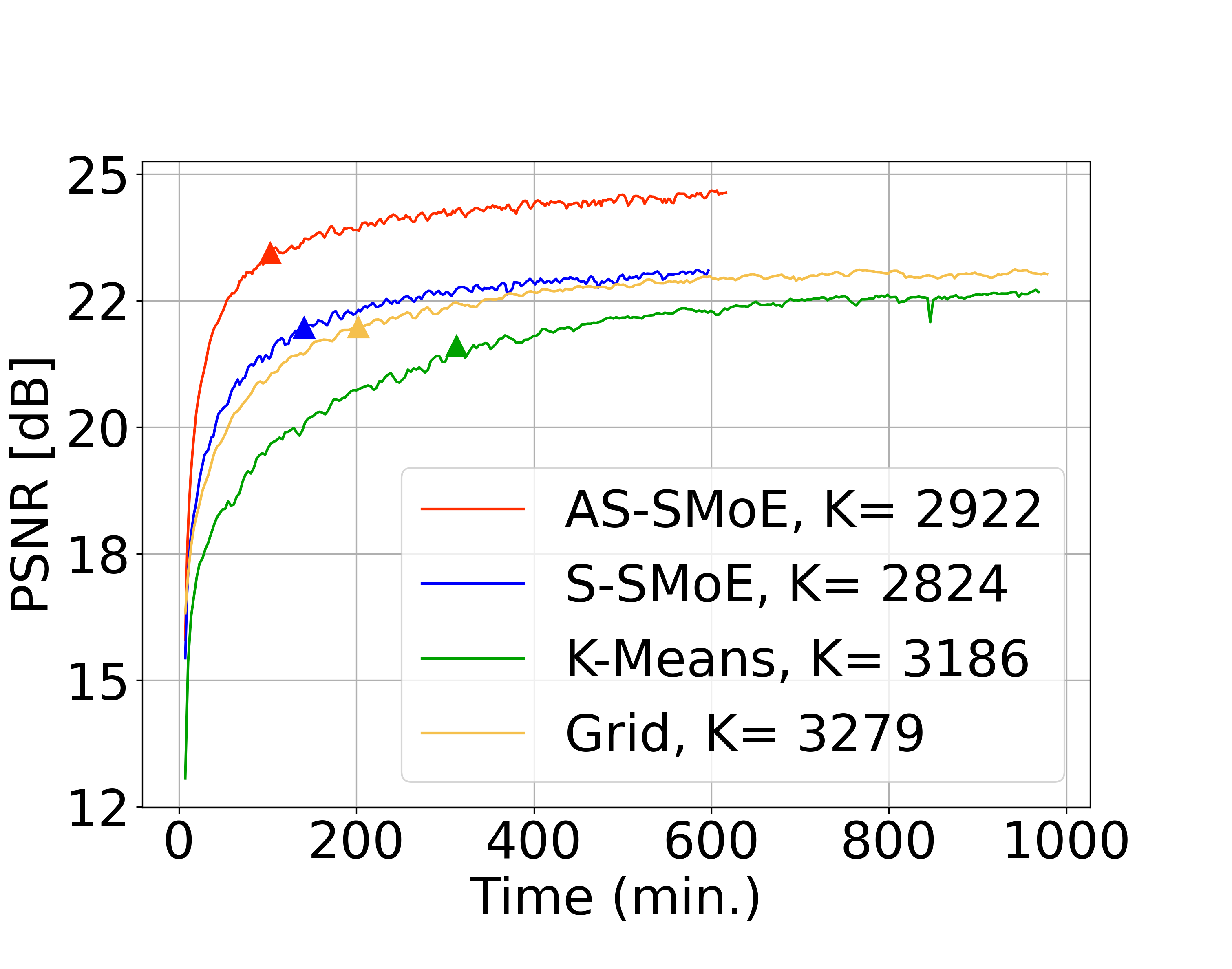}}
\subfloat[\textit{flower}]{\includegraphics[trim={0.1cm 1cm 2.2cm 2cm},clip,width=.33\linewidth]{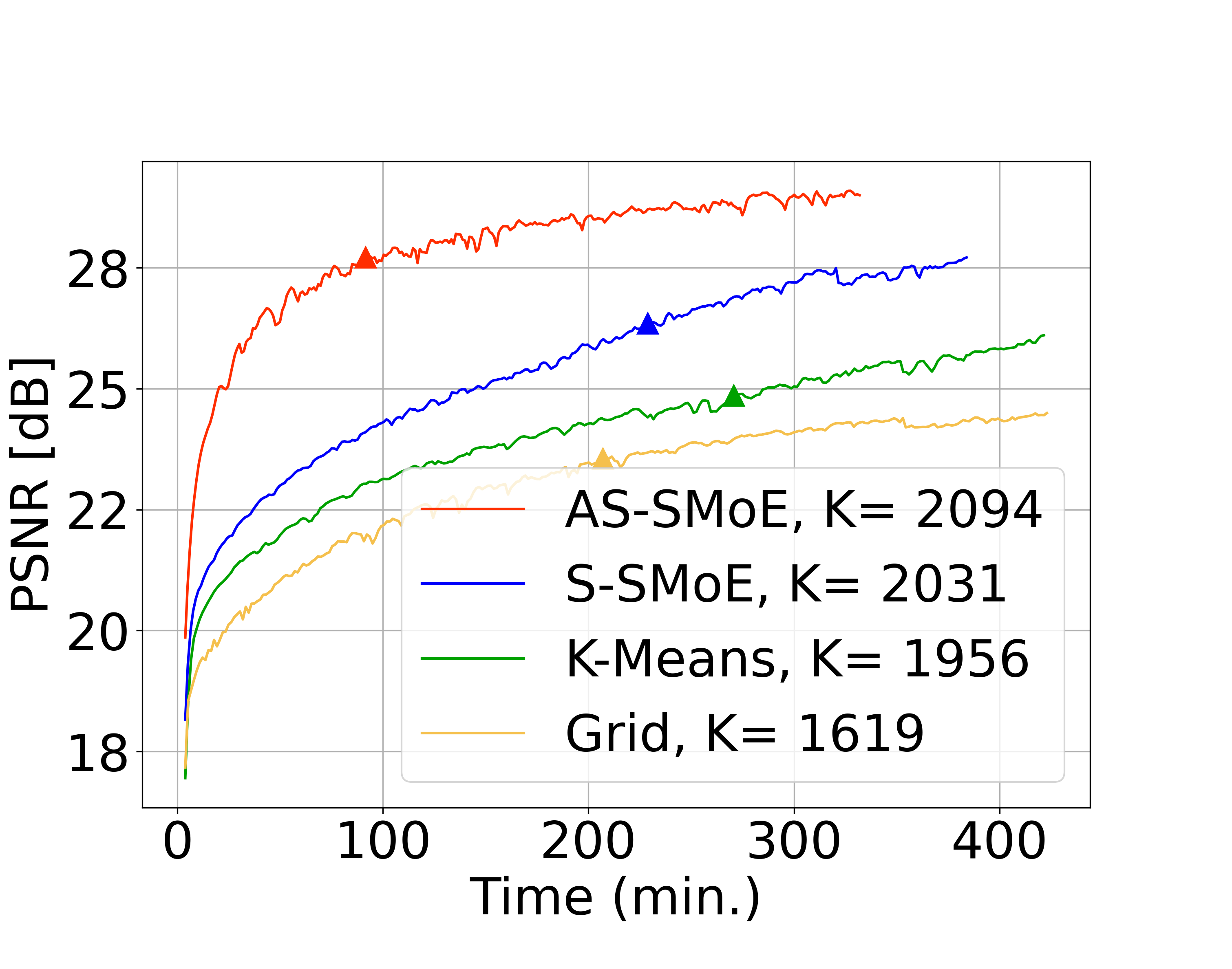}}
\subfloat[\textit{Run-time with different initialization}]{\includegraphics[trim={0.0cm 0cm 0cm 0cm},clip,width=.34\linewidth]{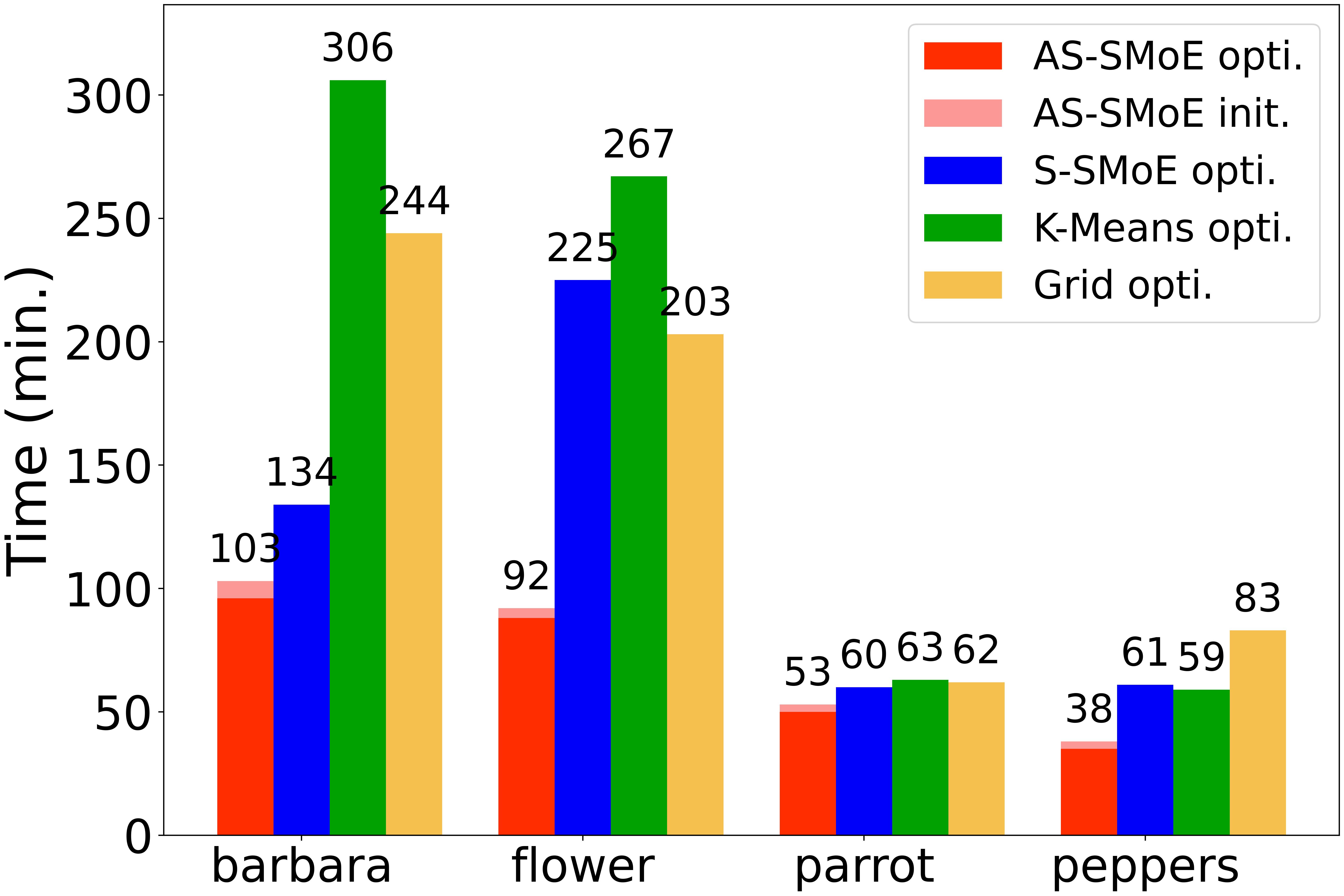}}

\caption{ (a) and (b) Convergence rate with Grid (yellow), K-Means (green), S-SMoE (blue), and AS-SMoE (red) initialization. The triangle markers indicate the time to achieve 95\% of the best PSNR performance. (c) Run-time for optimization of SMoE models with different initialization. For AS-SMoE, the run-time for initialization is depicted separately.}
\vspace{-.5em}
\label{fig9}
\end{figure*}

It is expected that improved initialization with better quality and fewer initial kernels $L_{Init}$ also reduces run-time of SMoE optimization. In each iteration the images need to be reconstructed from the SMoE model and the model dimension $L_{Init}$ impacts heavily on the run-time per iteration. On the other hand, better reconstruction is tuned to reduce the number of iterations resulting in reduced overall run-time to convergence. Fig.~\ref{fig9}a and b illustrate that AS-SMoE initialization in fact results in a much improved convergence rate. The target PSNR is individually set for each testing image, precisely at 0.95 times the best achievable results (triangle symbols). Fig.~\ref{fig9}c provides a comparative view of the time required by each method to achieve a predefined PSNR target after optimization. Results refer to the 0.95 triangular marked results in Fig.~\ref{fig9}a and b. The AS-SMoE initialization reduces the optimization run-time by great numbers (for images ``Barbara" and ``Flower" achieving a run-time reduction of 50\% from approximately 5 to 2.5~hrs and 4 to 2~hrs, respectively). For comparison, the run-time for the adaptive segmentation based initialization (AS-SMoE init) is also provided. This initialization is by far more complex than the Grid, K-Means, and S-SMoE initialization steps. The impact on the overall AS-SMoE run-time is noticeable, but does not increase the run-time significantly. This more complex initialization, and computationally more demanding, results in a superior quality, which drastically reduces run-time in succeeding global optimization. 

\begin{figure}[t]
\captionsetup{font=small}
\centerline{\includegraphics[width=3.0in]{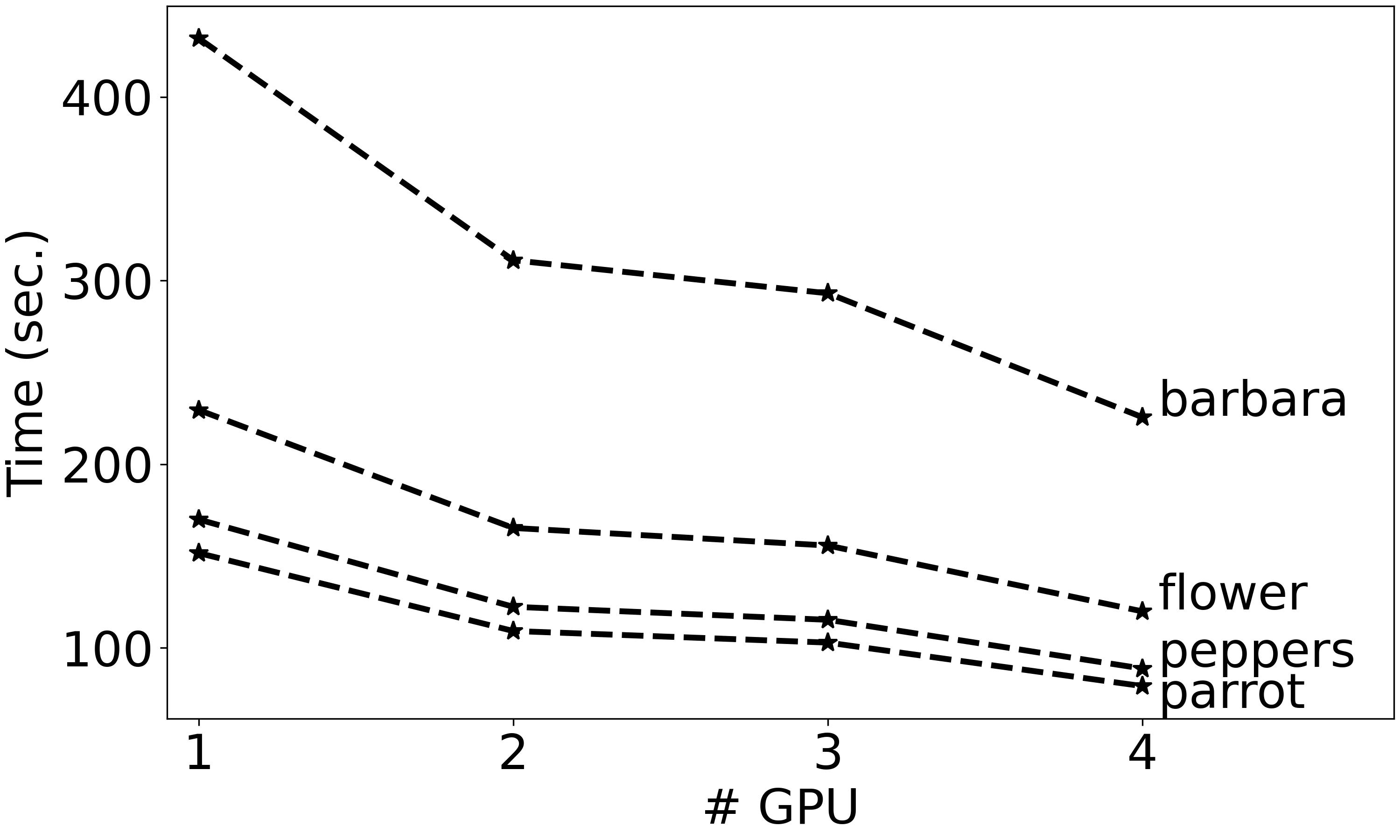}}
\vspace{-0.5em}
\caption{Initialization time of the proposed method depending on the number of GPUs.}
\vspace{-1em}
\label{fig10}
\end{figure}

The computational gains for optimizing SMoE models on single GPU carry over for steered RBF network Kernel Regression (AS-RBF), which is not shown in Fig.~\ref{fig9}c. As an important further advantage, AS-SMoE and AS-RBF initialization can be implemented with parallel computation to further reduce run-time. This is a result of the segmentation-based representation, where each segment can be processed and optimized individually. Fig.~\ref{fig10} depicts run-time saving of the initialization on multiple GPUs, with savings of approximately 50\% with 4 GPUs. In theory the processing can be divided into as many tasks as there are segments in an image, giving rise to further significant run-time savings.

\section{Summary and Conclusion}

In this paper, we introduced and investigated a novel adaptive segmentation-based initialization method to improve the optimization efficiency of kernel image regression methods. Particular emphasis was placed on optimizing Steered-Mixture-of Experts (SMoE) regression gating networks as well as Radial-Basis-Function (RBF) networks with steered Gaussian kernels. In contrast to many Deep Neural Networks used for similar processing tasks, ``shallow" SMoE and RBF networks allow a high interpretability of the regression results. We demonstrated that in SMoE networks, the impact of kernel parameters, expert values and gating functions on pixel reconstruction is directly encoded in the kernel parameters.

The adaptive segmentation-based initialization method leverages the interpretability of kernel-based models, efficiently allocating kernels to pre-calculated segments. Optimal kernel numbers, kernel positions, and steering parameters are derived per segment through iterative optimization and sparsification, enabling a ``global" initialization for SMoE, RBF, and similar kernel regression methods.

This approach yields significant improvements in both objective and subjective quality compared to ``state-of-the-art" K-Means initialization and our previously introduced segmentation-based initialization method, particularly enhancing the reconstruction of edges and lines. The novel initialization reduced model complexity and sparsity, crucial for applications like image compression and noise reduction. while also cutting the iterations and run-time for gradient descent optimization by up to 50\%.

Despite its complexity compared to K-Means, the novel method does not significantly impact the overall run-time but reduces the global optimization step. It allows heavy parallel computation, achieving a 50\% run-time reduction using four parallel GPUs.

Further refinement of the method through advanced segmentation algorithms and testing across various datasets and imaging scenarios will enhance its generalizability and robustness. Although primarily tested with SMoE and RBF networks with steering kernels, this initialization strategy is expected to benefit other image regression methods, including those with radial and Epanechnikov kernels \cite{liu_4d_2022}. Extending this framework to higher-dimensional data, like Magnetic Resonance Spectroscopy and light field data \cite{kerbl_3d_2023}, may yield even greater improvements.

\bibliographystyle{IEEEtran}
\bibliography{refs}

\vspace{-33pt}
\begin{IEEEbiographynophoto}{Yi-Hsin Li}
She started a double degree PhD program at Technical University Berlin, Germany and seconded to MIUN in November 2021. She in her four-year PhD journey works on high-dimensional data compression, focusing on gating networks. 
\end{IEEEbiographynophoto}
\vspace{-33pt}
\begin{IEEEbiographynophoto}{Sebastian~Knorr}
Sebastian Knorr, professor of Visual Computing at HTW Berlin. Broad competence in 3D/ 360°/ light field imaging technologies. Expertise in capture and synthesis of stereo 3D and 360° video; visual attention and quality assessment in VR; light field imaging and neural radiance fields. 
\end{IEEEbiographynophoto}
\vspace{-33pt}
\begin{IEEEbiographynophoto}{Mårten~Sjöström}
Mårten Sjöström, professor of signal processing, Mid Sweden University. Multidimensional signal processing, imaging and compression; system modelling and identification; inverse problems using machine learning. 120+ scientific articles, two book chapters. Supervisor of 8 PhD students, 8 previously.
\end{IEEEbiographynophoto}
\vspace{-33pt}
\begin{IEEEbiographynophoto}{Thomas~Sikora}
Thomas Sikora, professor and director of the Communication Systems Lab, Technische Universität Berlin, Germany.
\end{IEEEbiographynophoto}

\end{document}